\documentclass[10pt,twocolumn,letterpaper]{article}
\usepackage[table]{xcolor}
\usepackage{cvpr}              
\usepackage{float}

\usepackage[breakable]{tcolorbox}
\definecolor{cvprblue}{rgb}{0.21,0.49,0.74}
\usepackage[pagebackref,breaklinks,colorlinks,allcolors=cvprblue]{hyperref}
\usepackage{multirow}
\usepackage{arydshln}
\usepackage{enumitem}

\def\confName{CVPR}
\def\confYear{2026}

\title{MedGRPO: Multi-Task Reinforcement Learning for \\ Heterogeneous Medical Video Understanding} 

\author{
Yuhao Su$^{1,2,*}$\quad
Anwesa Choudhuri$^{2,\dagger}$\quad
Zhongpai Gao$^{2,\dagger}$\quad
Benjamin Planche$^{2}$\\
Van Nguyen Nguyen$^{2}$\quad
Meng Zheng$^{2}$\quad
Yuhan Shen$^{1}$\quad
Arun Innanje$^{2}$\\
Terrence Chen$^{2}$\quad
Ehsan Elhamifar$^{1, \ddagger}$\quad
Ziyan Wu$^{2, \ddagger}$\\
$^{1}$Northeastern University, Boston, MA, USA\\
$^{2}$United Imaging Intelligence, Boston, MA, USA\\
}
\begin{document}
\maketitle
\begingroup
\renewcommand\thefootnote{}\footnotetext{
$^{*}$This work was carried out during the internship of Yuhao Su at 
United Imaging Intelligence, Boston MA, USA.
$^{\dagger}$Corresponding authors. $^{\ddagger}$Co-last authors.
}
\endgroup

\begin{abstract}
Large vision-language models struggle with medical video understanding, where spatial precision, temporal reasoning, and clinical semantics are critical. To address this, we first introduce \textbf{MedVidBench}, a large-scale benchmark of 531,850 video-instruction pairs across 8 medical sources spanning video, segment, and frame-level tasks, curated through a rigorous quality assurance pipeline with expert-guided prompting and dual-model validation. While supervised fine-tuning on MedVidBench yields noticeable gains, standard Reinforcement Learning (RL) fails due to imbalanced reward scales across datasets, which destabilizes optimization and leads to training collapse. To overcome this, we introduce \textbf{MedGRPO}, a novel RL framework for balanced multi-dataset training with two key innovations: (1) \emph{cross-dataset reward normalization} that maps each dataset's median performance to a common reward value, ensuring fair optimization regardless of difficulty, and (2) a \emph{medical LLM judge} that evaluates caption quality on five clinical dimensions through comparative similarity scoring. 
Supervised fine-tuning Qwen2.5-VL-7B on MedVidBench outperforms GPT-4.1 and Gemini-2.5-Flash across all tasks, while MedGRPO further improves the SFT baseline on grounding and captioning. Our work establishes a foundational benchmark and training methodology for advancing medical video understanding with VLMs. Our project website is available at: \url{https://uii-america.github.io/MedGRPO/}.
\end{abstract}
\begin{figure*}[t]
\centering
\includegraphics[width=1\linewidth, height=8.6cm]{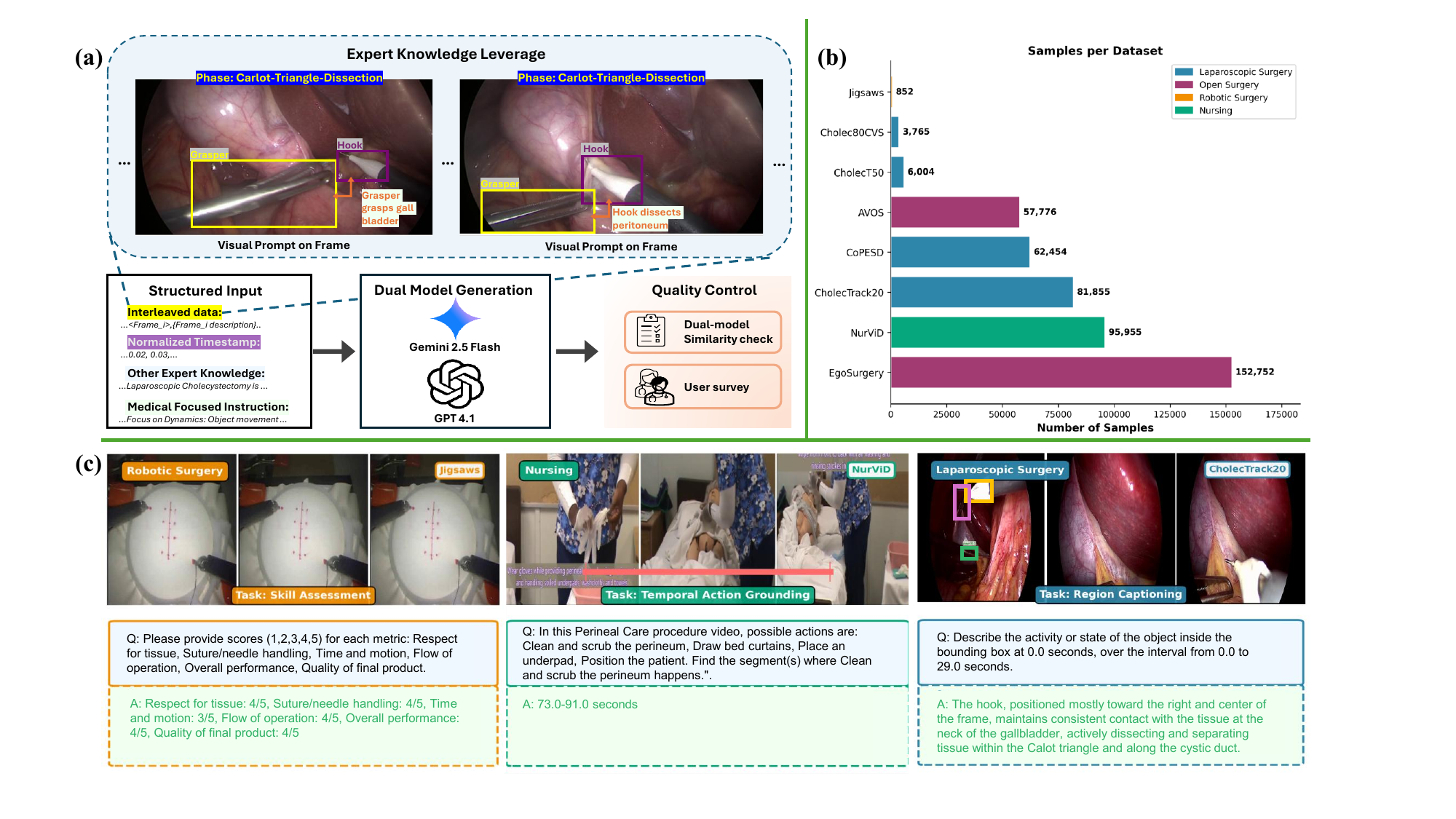}
\caption{Overview of MedVidBench. (a) High quality data curation pipeline for MedVidBench. We leaverage expert knowledge into prompt construction and generate high quality text using 2 VLMs (Gemini-2.5-Flash and GPT-4.1). (b) MedVidBench comprises of 8 different datasets, with 532k samples in total, spanning 4 different domains. (c) Examples of diverse tasks across different domains.}
\label{fig:dataset_overview}
\vspace{-1.5em}
\end{figure*}
\vspace{-0.5em}
\section{Introduction}
Large vision-language models (VLMs) have demonstrated remarkable capabilities in understanding and reasoning about general-domain visual content~\cite{flamingo,blip2,llava,qwen2vl}. However, their performance significantly degrades when applied to high-stake, expert-driven domains such as medicine. Medical video understanding presents unique challenges beyond general video understanding problems such as temporal modeling~\cite{su2024twostage} and object grounding~\cite{Su_2026_WACV}: it requires interpreting fine-grained surgical actions, understanding domain-specific terminology (e.g., correctly identifying a ``grasper'' rather than generic ``tool''), assessing procedural safety, and reasoning about multi-phase temporal workflows—requirements far beyond the everyday activities on which most models are trained.

Current state-of-the-art VLMs, including GPT-4.1~\cite{gpt4v}, Gemini-2.5-Pro~\cite{gemini25}, and Qwen2.5-VL~\cite{qwen2.5vl}, struggle to provide adequate performance on medical video tasks. The primary bottleneck is the lack of suitable training data in instruction-following format. While existing medical video datasets such as CholecT50~\cite{cholect50}, EgoSurgery~\cite{egosurgery}, AVOS~\cite{avos}, and NurViD~\cite{nurvid} contain rich annotations, ranging from frame-wise bounding boxes, action triplets to phase labels, they are not designed for VideoLLM which requires question-answer pairs in conversational format. Recent efforts like SurgLaVi~\cite{surglavi} and SurgLLM~\cite{surgllm} remain limited in task diversity and primarily focus on surgical procedures. Converting existing medical video annotations into high-quality instruction-following QA pairs at scale presents a fundamental challenge: medical annotations require expert-level understanding to transform into natural language that preserves clinical accuracy.

We introduce MedVidBench, a large-scale instructional dataset comprising 532K video-instruction pairs across 8 medical sources and 8 diverse tasks spanning three temporal granularities: video-level understanding (summarization, critical view of safety, next action prediction, skill assessment), segment-level reasoning (temporal action grounding, dense captioning, region captioning), and frame-level grounding (spatiotemporal localization). The core innovation lies in our multi-perspective quality assurance pipeline that systematically converts existing expert annotations into instruction-following format. Rather than creating annotations from scratch, we leverage GPT-4.1 and Gemini-2.5-Flash to transform existing annotations—frame-wise bounding boxes, action triplets, phase labels—into natural language QA pairs. Our pipeline employs source-specific strategies: for datasets with dense frame-wise annotations (CholecT50, EgoSurgery), we use visual prompting by overlaying bounding boxes and labels on frames; for web-sourced videos (AVOS, NurViD), we extract high-quality transcripts using Whisper-X~\cite{whisperx} and enrich prompts with video metadata (e.g., video titles). Dual-model validation ensures consistency and reduces generation errors.

While Supervised Fine-Tuning (SFT) a VLM on MedVidBench establishes a strong baseline, naively applying standard RL algorithms, such as GRPO~\cite{deepseekmath}, to our heterogeneous dataset to improve performance reveals a critical failure mode: training collapse. Models rapidly overfit to easy datasets—for instance, CoPESD~\cite{copesd} achieves median spatiotemporal grounding mIoU around 0.5—while performance on challenging datasets like EgoSurgery~\cite{egosurgery} (median mIoU around 0.12) degrades dramatically. This occurs because raw task metrics used as rewards create fundamentally unfair optimization: the model receives consistently higher rewards for easy dataset samples, causing gradient updates to prioritize easy sources while destabilizing learning on hard ones. A second challenge emerges in caption generation tasks: standard semantic similarity metrics fail to capture medical correctness. Two captions may achieve high surface-level similarity yet differ critically in instrument specificity (``tool'' vs. ``grasper''), action precision (``grasps'' vs. ``dissects''), anatomical accuracy (``tissue'' vs. ``cystic duct''), and spatial detail (``upper area'' vs. ``upper right quadrant'').

We introduce MedGRPO, a reinforcement learning framework that addresses these challenges through two key innovations. First, cross-dataset reward normalization uses logistic functions centered on dataset-specific percentile statistics to map each dataset's median performance to a fixed reward value, ensuring balanced optimization regardless of task difficulty. This normalization provides smooth gradients while being robust to outliers. Second, for caption generation tasks, we design a medical LLM judge that uses comparative similarity scoring to evaluate how closely generated captions match references across five clinical dimensions: medical terminology precision, instrument and anatomy identification, specificity versus vagueness, clinical procedure context, and action accuracy. This comparative framing avoids score inflation from absolute quality ratings. For reward computation during training, we combine LLM judge scores with semantic similarity in a hybrid design, leveraging both fine-grained clinical correctness and overall semantic coherence.

We first train Qwen2.5-VL-7B~\cite{qwen2.5vl} using supervised fine-tuning (SFT) on MedVidBench and outperform closed-source alternatives including GPT-4.1 and Gemini-2.5-Flash. We then use reinforcement learning with MedGRPO to train this SFT baseline and achieve substantial improvements over the SFT baseline across nearly all tasks while maintaining balanced learning across both easy and hard datasets. Our main contributions are:
\begin{itemize}[leftmargin=*,itemsep=1pt,topsep=2pt]
    \item \textbf{MedVidBench}: A benchmark of 532K video-instruction pairs across 8 medical sources and 8 tasks spanning video/segment/frame-level understanding, curated through a multi-perspective quality assurance pipeline with source-specific strategies and dual-model validation.
    \item \textbf{MedGRPO}: A reinforcement learning framework enabling balanced multi-dataset training through (1) cross-dataset reward normalization using logistic functions for median-centered fairness, and (2) a medical LLM judge evaluating five clinical dimensions via  by comparing how closely generated captions match references.
    \item \textbf{Comprehensive evaluation} demonstrating balanced improvements across heterogeneous datasets, superiority over closed-source models (GPT-4.1, Gemini-2.5-Flash), and ablations confirming that without reward normalization, training collapses with increasing entropy.
\end{itemize}

\begin{figure*}[t]
\centering
\includegraphics[width=\linewidth,height=4.5cm]{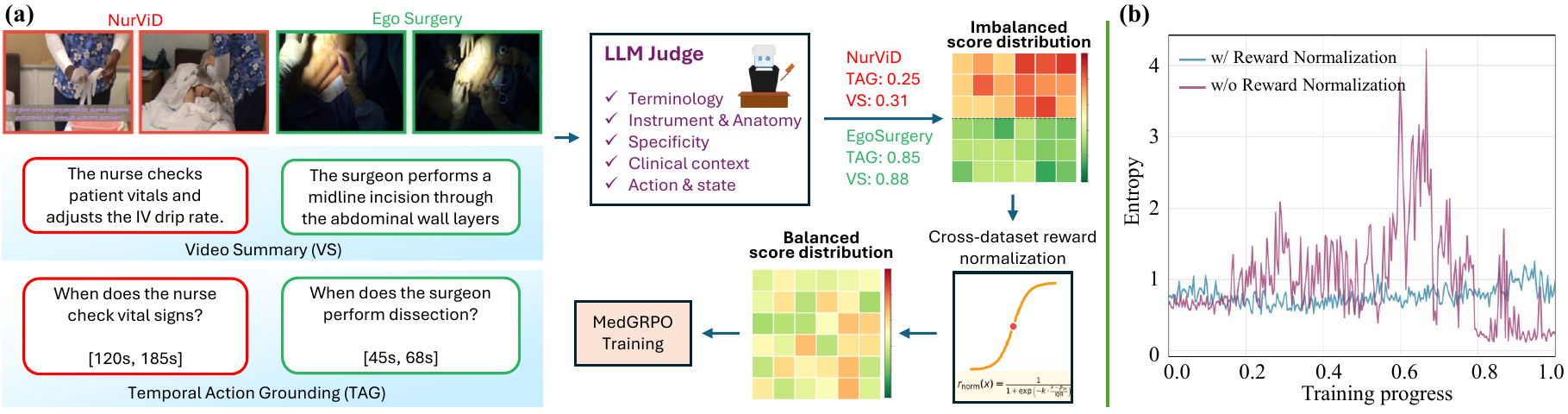}
\caption{Overview of MedGRPO. (a) MedGRPO framework with cross-dataset reward normalization and medical LLM judge evaluation. (b) Training entropy comparison between models trained with and without reward normalization.}
\label{fig:method_overview}
\vspace{-1.5em}
\end{figure*}

\section{Related Work}
\label{sec:related}

\noindent\textbf{Medical Video Datasets.} Medical video understanding relies on specialized datasets capturing clinical procedures. Surgical datasets like CholecT50~\cite{cholect50} and EgoSurgery~\cite{egosurgery} provide frame-level annotations including action triplets and phase labels, but are limited in scale and diversity. Web-sourced datasets like AVOS~\cite{avos} and NurViD~\cite{nurvid} offer greater diversity with phase labels from which we extract audio, but lack frame-level precision. However, existing datasets provide annotations for traditional computer vision tasks rather than conversational question-answering format required by VideoLLMs, and remain isolated by procedure type. Our work systematically converts existing expert annotations into instruction-following format and enables joint training across 8 heterogeneous sources through reward normalization.

\vspace{0.4em}
\noindent\textbf{General Video LLMs.} Large vision-language models have demonstrated impressive capabilities in general video understanding. Early frameworks like Flamingo~\cite{flamingo} and BLIP-2~\cite{blip2} established cross-modal alignment on images, with recent VideoLLMs like Video-ChatGPT~\cite{videochatgpt}, VideoLLaMA~\cite{videollama3}, and Video-LLaVA~\cite{videollava} extending these to videos through temporal modeling and video instruction tuning. State-of-the-art models including GPT-4.1~\cite{gpt4v}, Gemini-2.5-Flash~\cite{gemini25}, and Qwen2.5-VL~\cite{qwen2.5vl} demonstrate strong performance on general-domain benchmarks~\cite{activitynet,msrvtt}. However, medical video understanding presents unique challenges: precise instrument identification, fine-grained action recognition, anatomical structure localization, and temporal reasoning about multi-step procedures.

\vspace{0.4em}
\noindent\textbf{Medical VideoLLMs.} Medical video understanding requires specialized knowledge beyond general vision-language models. Early medical VLMs focused on static radiology images (MedVInT~\cite{medvint}, LLaVA-Med~\cite{llavamed}, Med-Flamingo~\cite{medflamingo}), while recent work addresses RGB surgical videos with temporal dynamics. SurgLaVi~\cite{surglavi} introduces 240K surgical clip-caption pairs for CLIP-style contrastive pretraining but remains in descriptive format rather than instruction-following. Building on this, SurgLLM~\cite{surgllm} fine-tunes Qwen2-VL on surgical instruction data from Cholec80~\cite{cholec80}, a laparoscopic cholecystectomy dataset, demonstrating improved phase recognition and captioning. Similarly, Qwen2.5VL-7B$_{\text{Surg-CholecT50}}$~\cite{nvidia2025surgqwen} fine-tunes on CholecT50 for surgical workflow analyses. However, these approaches rely on single-dataset training, limiting cross-procedure generalization and task coverage, with no mechanisms to handle difficulty disparity across heterogeneous medical video datasets.
Concurrent efforts advance medical AI through multimodal reasoning, surgical foundation models, and knowledge distillation~\cite{gong2025med,che2025lemon,che2025stitch,Lan_2025_ICCV,lan2026reco,lan2026performance,su2025streamline,guo2026momentum}, yet multi-task learning in medical video understanding remains unexplored. Our work addresses this with a unified framework across 8 datasets and 8 tasks, enabling balanced learning via cross-dataset reward normalization.


\vspace{0.4em}
\noindent\textbf{Multi-Task Learning in VLMs.} Multi-task learning leverages task synergies to improve generalization in vision-language models. Works like Unified-IO~\cite{unifiedio} and OFA~\cite{ofa} demonstrate that single transformers can handle diverse modalities through task-specific prompting. Multi-dataset training introduces optimization challenges including catastrophic forgetting~\cite{gem,packnet} and requires careful gradient management. Recent work applies RL to align VLMs with preferences~\cite{rlhfv,silkie}, but focuses on single-domain or single-task settings. Existing approaches either require architectural modifications (separate task heads~\cite{unifiedio}, task-specific encoders~\cite{packnet}, or task prompts~\cite{ofa}) or assume homogeneous difficulty distributions. In medical video understanding, no prior work addresses the challenge we identify: when datasets have vastly different difficulty (e.g., median performance 0.5 vs 0.12), standard RL causes training collapse due to reward magnitude disparities. Our cross-dataset reward normalization provides a simple yet effective solution without architectural changes, enabling balanced multi-dataset learning.

\vspace{0.4em}
\noindent\textbf{Evaluation Metrics for Video Captioning.} Automatic evaluation of video captioning has evolved from n-gram metrics~\cite{bleu,meteor,cider} to embedding-based approaches like BERTScore~\cite{bertscore} and CLIPScore~\cite{clipscore}. 
Recent LLM-as-judge methods~\cite{geval,gptscore,prometheus,gavie,lee2024vhelm} show strong correlation with human judgment, including in clinical settings~\cite{croxford2025evaluating}. However, medical video evaluation remains underexplored, and standard metrics fail to distinguish critical clinical differences like instrument specificity (`tool' vs. `grasper'), action precision (`grasps' vs. `dissects'), and anatomical accuracy (`tissue' vs. `cystic duct'). We introduce a medical LLM judge using comparative evaluation to assess caption quality on five clinical dimensions, combined with semantic similarity for robust RL training.

\section{Method}
\label{sec:method}
We begin by introducing MedVidBench (\S\ref{sec:dataset}), our large-scale benchmark created through a multi-perspective quality assurance pipeline (Figure~\ref{fig:dataset_overview}). We then present MedGRPO (Figure~\ref{fig:method_overview}), a reinforcement learning framework designed for fair multi-dataset training on MedVidBench: two-stage training paradigm (\S\ref{sec:training}) and two core technical innovations of cross-dataset reward normalization (\S\ref{sec:reward_norm}) that prevents catastrophic forgetting across heterogeneous datasets, and a medical LLM judge (\S\ref{sec:llm_judge}) that captures domain-specific correctness beyond surface metrics.
\subsection{MedVidBench: Multi-Granular Medical Video Benchmark}
\label{sec:dataset}

We construct MedVidBench, a unified benchmark spanning 8 medical video sources across 8 tasks with 531,850 video-instruction pairs in conversational QA format (Figure~\ref{fig:dataset_overview},Table~\ref{tab:dataset_comparison}). Our benchmark systematically transforms existing expert annotations---bounding boxes, procedure transcripts, action labels---into instruction-following format through a multi-perspective quality assurance pipeline with dual-model validation.

\begin{table}[t]
\centering
\caption{Comparison with existing medical instruction datasets. Our MedVidBench provides superior scale, task coverage, multi-domain annotations, and quality assurance.}
\setlength{\tabcolsep}{2.5pt}
\footnotesize
\resizebox{\linewidth}{!}{
\begin{tabular}{l|cccrc}
\toprule
\textbf{Dataset} & \textbf{Format} & \textbf{Domain} & \textbf{Tasks} & \textbf{Scale} & \textbf{Quality Assurance} \\
\midrule
SurgLaVi~\cite{surglavi}  & CLIP-style & Surgery & 1 & 240K & Dual-model \\
Surg-396K~\cite{endochat} & Image-QA & Surgery & 5 & 396K & Single model \\
SVU-31K~\cite{surgvidlm}  & Video-QA & Surgery & 3 & 31K & Single model \\
\midrule
\textbf{MedVidBench} & \multirow{2}{*}{\textbf{Video-QA}} & \textbf{Surgery} & \multirow{2}{*}{\textbf{8}} & \multirow{2}{*}{\textbf{532K}} & \textbf{Dual-model} \\
\textbf{(Ours)} &  & \textbf{\& Instructional} &  &  & \textbf{validation} \\
\bottomrule
\end{tabular}}
\label{tab:dataset_comparison}
\vspace{-1em}
\end{table}

\noindent\textbf{Data Sources and Task Coverage.}
MedVidBench integrates 8 diverse medical video sources 
covering laparoscopic surgery (CholecT50~\cite{cholect50}, CholecTrack20~\cite{cholectrack20}, Cholec80-CVS~\cite{cholec80cvs}, CoPESD~\cite{copesd}), open surgery (AVOS~\cite{avos}, EgoSurgery~\cite{egosurgery}), robotic surgery (JIGSAWS~\cite{JIGSAWS}), and nursing procedures (NurViD~\cite{nurvid}). Our benchmark spans 626 unique videos with duration range 20s--1800s and adaptive FPS 0.1--3.0 optimized per task. We design tasks across three temporal granularities: 1) \textit{Video-level tasks} include Video Summarization (\texttt{VS}), Critical View of Safety (\texttt{CVS}), Next Action Prediction (\texttt{NAP}), and Skill Assessment (\texttt{SA}), requiring holistic understanding of entire procedures; 2) \textit{Segment-level tasks} include Temporal Action Grounding (\texttt{TAG}), Dense Video Captioning (\texttt{DVC}), and Region Captioning (\texttt{RC}), enabling event-level reasoning; 3) \textit{Frame-level tasks} include Spatiotemporal Grounding (\texttt{STG}) for precise instrument localization. This multi-granular design enables models to understand \emph{when} actions occur (temporal), \emph{where} instruments are located (spatial), and \emph{how} procedures should be performed (procedural context).

\vspace{0.4em}
\noindent\textbf{Multi-Perspective Quality Assurance Pipeline.}
Transforming existing medical annotations into instruction-following format requires expert-level precision (e.g., distinguishing ``Maryland dissector'' from generic ``tool''). We develop a three-stage pipeline: (1) \textit{Expert annotation prompting:} leverage dataset-specific strategies---for frame-annotated datasets (e.g., CholecT50, EgoSurgery), we overlay bounding boxes and labels directly on frames with procedure-specific context; for web-sourced datasets (AVOS, NurViD), we extract transcripts from high-quality audio using Whisper-X~\cite{whisperx} and enrich with metadata; (2) \textit{Dual-model generation:} generate captions independently with GPT-4.1 and Gemini-2.5-Flash, using both models for video summary and dense captioning tasks to prevent model-specific biases; (3) \textit{Quality validation:} compute caption similarities between GPT-4.1 and Gemini-2.5-Flash outputs using sentence-transformers~\cite{sentencebert}, filter low-quality pairs (similarity $<$0.3), sample videos with 50--180 frames using adaptive FPS (0.1--3.0) suitable for different task requirements, and create video-level train/test splits (test ratio 0.15). This rigorous pipeline produces 531,850 high-quality instances (461,413 training, 70,437 test).

\vspace{0.4em}
\noindent\textbf{Human Validation Study.} We conduct user studies with medical professionals to empirically validate our expert annotation prompting approach. Medical experts compare captions generated using our annotation-enriched prompts (with overlaid bounding boxes, procedure context, and enriched transcripts) versus baseline generation from raw video frames only, rating on clinical accuracy and terminology precision. Results demonstrate significant preference for annotation-enriched generation, confirming that expert annotation prompting produces superior medical video QA pairs compared to naive frame-based generation. More dataset statistics, data curation details, LLM-judge prompts, and human study detail in the supplementary.

\subsection{MedGRPO Training Paradigm}
\label{sec:training}
Our goal is to build a multi-task medical VideoLLM handling diverse tasks across heterogeneous datasets. We employ a two-stage approach: supervised fine-tuning (SFT) first adapts Qwen2.5-VL-7B to medical video understanding, injecting domain-specific knowledge while preserving the pretrained model's general capabilities; this establishes baseline performance for percentile computation. Subsequently, reinforcement learning with GRPO enables multi-task improvement by aligning outputs with medical expertise through cross-dataset reward normalization (§\ref{sec:reward_norm}) and medical-specific LLM-judge evaluation (§\ref{sec:llm_judge}), without modifying the model architecture.

We apply GRPO~\cite{deepseekmath}, a policy gradient method that avoids value function training and naturally handles diverse reward scales through group-relative advantage estimation. For a given prompt $q$, we sample a group of $G=8$ responses $\{o_i\}_{i=1}^G$ from our current policy $\pi_\theta$. We compute the advantage $\hat{A}_{i}$ for each response $o_i$ by normalizing the response's reward $r_i$ within each group:
\begin{equation}
\hat{A}_{i} = \frac{r_i - \text{mean}(\{r_j\}_{j=1}^G)}{\text{std}(\{r_j\}_{j=1}^G)}
\end{equation}
The policy $\pi_\theta$ is optimized using the clipped surrogate objective against $\pi_{\theta_{\text{old}}}$ (the policy before the gradient update):
\begin{equation}
\small
\mathcal{L}_{\text{GRPO}} = \mathbb{E}\left[\sum_t \min\left(\rho_t \hat{A}_{i}, \text{clip}(\rho_t, 1-\epsilon_\text{low}, 1+\epsilon_\text{high}) \hat{A}_{i}\right)\right]
\end{equation}
where $\rho_t = \pi_\theta(o_{i,t}|q,o_{i,<t}) / \pi_{\theta_{\text{old}}}(o_{i,t}|q,o_{i,<t})$ is the importance sampling ratio at token position $t$. Following DAPO~\cite{dapo}, we adopt asymmetric clipping with $\epsilon_\text{low} = 0.2$ and $\epsilon_\text{high} = 0.3$ to allow larger positive updates while constraining negative ones, and remove the KL penalty term from standard GRPO for improved performance.
\subsection{Cross-Dataset Reward Normalization}
\label{sec:reward_norm}

To enable fair multi-dataset multi-task optimization, we normalize rewards using dataset-task-specific logistic functions. While mapping rewards to $[0,1]$ is common in GRPO, our key contribution is \emph{stretching reward distributions} to have comparable spread across heterogeneous metrics. This addresses two sources of imbalance: datasets exhibit vastly different difficulty levels (e.g., CoPESD STG with median performance 0.5 vs. EgoSurgery STG at 0.12), and tasks use incomparable metrics (e.g., mIoU for grounding vs. our medical LLM judge scores (§\ref{sec:llm_judge}) for captioning). The key insight is \emph{median fairness}: median-level performance receives equal normalized rewards across all dataset-task pairs, eliminating bias in gradient updates.

\vspace{0.4em}
\noindent\textbf{Normalization Function.}
For each dataset-task combination $(d, t)$, we apply logistic transformation to metric $x$:
\begin{equation}
r_{\text{norm}}^{(d,t)}(x) = \frac{1}{1 + \exp\left(-k \cdot \frac{x - p_{50}^{(d,t)}}{\text{IQR}^{(d,t)}}\right)}
\label{eq:logistic_norm}
\end{equation}
where $p_{50}^{(d,t)}$ is the median, $\text{IQR}^{(d,t)} = p_{75}^{(d,t)} - p_{25}^{(d,t)}$ is the interquartile range, and $k=3.0$ controls the sigmoid slope. The percentile statistics $\{p_{25}, p_{50}, p_{75}\}$ are computed from SFT baseline predictions on the training set for each dataset-task combination.

\vspace{0.4em}
\noindent\textbf{Design Properties.} This design provides four advantages: (1) \textit{Median fairness}---when $x = p_{50}^{(d,t)}$, the exponent becomes zero, yielding $r_{\text{norm}}^{(d,t)} = 0.5$ for all dataset-task pairs. For example, easy dataset-task pairs (CoPESD \texttt{STG}: $p_{50}\approx0.5$) and hard pairs (EgoSurgery \texttt{STG}: $p_{50}\approx0.12$) receive identical rewards at their respective medians, eliminating bias across both datasets and tasks. (2) \textit{Smooth gradients}---the logistic function provides non-zero derivatives everywhere: $\frac{dr_{\text{norm}}}{dx} = \frac{k}{\text{IQR}} \cdot r_{\text{norm}}(1 - r_{\text{norm}})$, avoiding dead zones from hard clipping. (3) \textit{Outlier robustness}---IQR-based scaling uses the middle 50\% of the distribution, unlike min-max normalization sensitive to range extremes. (4) \textit{Bounded output}---the logistic function maps to $(0, 1)$, compatible with GRPO's group normalization, unlike unbounded z-score normalization.

\vspace{0.4em}
\noindent\textbf{Task-Specific Reward Design.} We apply GRPO training to four representative tasks spanning different temporal granularities:  \texttt{VS} (video-level), \texttt{TAG} (segment-level), \texttt{RC} (segment-level) and \texttt{STG} (frame-level). For grounding tasks (\texttt{TAG}, \texttt{STG}), we use multiplicative composite rewards as penalties rather than additive bonuses, since the model reliably outputs valid structures after SFT: $r_{\text{temporal}} = r_{\text{content}} \times r_{\text{format}}$, where $r_{\text{content}}$ is logistic-normalized mIoU and $r_{\text{format}} = 1.0 - 0.6 \times (1 - \mathbb{I}_{\text{valid}})$ penalizes parsing failures. For captioning tasks (\texttt{VS}, \texttt{RC}), we combine semantic similarity and LLM judge evaluation (§\ref{sec:llm_judge}).

\subsection{Medical LLM Judge for Evaluation}
\label{sec:llm_judge}

For caption generation tasks (\texttt{VS}, \texttt{DVC}, \texttt{RC}), standard embedding-based metrics (SentenceBERT~\cite{sentencebert}, BERTScore~\cite{bertscore}) capture overall paragraph-level semantic similarity but fail to assess fine-grained medical correctness. For example, captions \textit{``The tool grasps tissue in the upper area"} vs. \textit{``The grasper dissects the cystic duct in the upper right quadrant"} achieve cosine similarity $\approx 0.82$ yet differ critically in \textit{instrument specificity} (generic ``tool" vs precise ``grasper"), \textit{action accuracy} (``grasps" vs ``dissects"), \textit{anatomical precision} (vague ``tissue" vs specific ``cystic duct"), and \textit{spatial detail} (``upper area" vs ``upper right quadrant"). Such distinctions are clinically significant but invisible to general-domain embeddings.

\begin{table*}[t]
\centering
\footnotesize
\caption{Main results on MedVidBench across 8 tasks. We compare off-the-shelf baselines from 2025 and 2026 (evaluated with one-shot prompting), our SFT baselines, and our full MedGRPO method on Qwen2.5VL-7B, Qwen3-VL-4B, and Qwen3.5-4B. We use accuracy for \texttt{CVS}/\texttt{NAP}/\texttt{SA}, mIoU for \texttt{STG}/\texttt{TAG}, LLM judge scores for \texttt{DVC}/\texttt{VS}/\texttt{RC}, and F1 score for \texttt{DVC} as metrics. The best scores within each group are highlighted with red and the second best with orange. MedGRPO results for Qwen3.5-4B will be added in a future update.}
\label{tab:main_results}
\resizebox{\linewidth}{!}{
\begin{tabular}{@{}lcccccccccc@{}}
\toprule
Model           & \texttt{CVS}$_{acc}$ & \texttt{NAP}$_{acc}$& \texttt{SA}$_{acc}$ & \texttt{STG}$_{mIoU}$ & \texttt{TAG}$_{mIoU@0.3}$ & \texttt{TAG}$_{mIoU@0.5}$ & $\texttt{DVC}_{F1}$ & \texttt{DVC}$_{llm}$ & \texttt{VS}$_{llm}$ & \texttt{RC}$_{llm}$ \\ \midrule
\multicolumn{11}{l}{\texttt{2025 Off-the-shelf Baselines}} \\
GPT-4.1 \cite{gpt4v}           & 0.018         & 0.250          & 0.087   & 0.014         & 0.096         & 0.005         &0.101          & 2.438           & 2.490          & 2.080          \\
Gemini-2.5-flash \cite{gemini25}    & 0.101         & 0.228            & 0.107      &  0.047         &  0.045         & 0.021         & 0.084          & 2.387           & 2.352          & 1.912          \\
VideoChat-R1.5-7B \cite{yan2025videochat}                    & 0.000    & 0.270    & 0.006   & 0.000     & 0.009         & 0.005         & 0.026   & 1.723    & 3.034   &  3.086   \\
\midrule
\multicolumn{11}{l}{\texttt{2025 Qwen2.5VL-7B}} \\
Qwen2.5VL-7B \cite{qwen2.5vl}      &  0.105         & 0.151       & 0.010    & 0.020            & 0.006         & 0.068         & 0.075          & 2.512           & 2.452          & 2.090          \\
Qwen2.5VL-7B$_{\text{Surg-CholecT50}}$ (NVIDIA) \cite{nvidia2025surgqwen} & 0.000    & 0.302    & 0.000   & 0.000     & 0.019         & 0.013         & 0.051   & 1.945    & 2.101   & \cellcolor{orange!25} 2.986   \\
Qwen2.5VL-7B$_{\text{SFT}}$ (\textbf{Ours})   & \cellcolor{orange!25} 0.894         & \cellcolor{red!25} \textbf{0.442}             &\cellcolor{orange!25} 0.218          
&\cellcolor{orange!25} 0.177  
& \cellcolor{orange!25}0.142         &\cellcolor{orange!25} 0.091         &\cellcolor{orange!25} 0.165          &\cellcolor{orange!25} 3.665           &\cellcolor{orange!25} 3.596          & 2.757          \\
Qwen2.5VL-7B$_{\text{MedGRPO}}$ (\textbf{Ours})      & \cellcolor{red!25} \textbf{0.896}         
& \cellcolor{orange!25} 0.405             
& \cellcolor{red!25} \textbf{0.254}          
& \cellcolor{red!25} \textbf{0.202}   
& \cellcolor{red!25} \textbf{0.216}         & \cellcolor{red!25} \textbf{0.156}         & \cellcolor{red!25} \textbf{0.214}          & \cellcolor{red!25} \textbf{3.797}           & \cellcolor{red!25} \textbf{4.184}          & \cellcolor{red!25} \textbf{3.442}          \\
\midrule
\multicolumn{11}{l}{\texttt{2025 Qwen3-VL-4B}} \\
Qwen3VL-4B \cite{qwen3vl}                          & 0.000    & 0.178    & 0.006   & 0.000     & 0.039         & 0.034         & 0.128   & 1.939    & 2.926   & 2.853   \\
Qwen3VL-4B$_{\text{SFT}}$ (\textbf{Ours})               & \cellcolor{orange!25} 0.895    & \cellcolor{orange!25} 0.466    & \cellcolor{orange!25} 0.270   & \cellcolor{orange!25} 0.133     & \cellcolor{orange!25} 0.465         & \cellcolor{orange!25} 0.403         & \cellcolor{orange!25} 0.435   & \cellcolor{orange!25} 3.862    & \cellcolor{orange!25} 4.180   & \cellcolor{orange!25} 3.752   \\
Qwen3VL-4B$_{\text{MedGRPO}}$ (\textbf{Ours})           & \cellcolor{red!25} \textbf{0.898}    & \cellcolor{red!25} \textbf{0.473}    & \cellcolor{red!25} \textbf{0.285}   & \cellcolor{red!25} \textbf{0.176}     & \cellcolor{red!25} \textbf{0.504}         & \cellcolor{red!25} \textbf{0.441}         & \cellcolor{red!25} \textbf{0.480}   & \cellcolor{red!25} \textbf{3.950}    & \cellcolor{red!25} \textbf{4.227}   & \cellcolor{red!25} \textbf{3.861}  \\
\midrule
\multicolumn{11}{l}{\texttt{2026 Off-the-shelf Baselines \& Qwen3.5-4B}} \\
GPT-5.4 \cite{gpt4v}           & 0.164         & 0.393          & 0.267   & 0.004         & 0.086         & 0.055         & 0.178          & 3.403           & 3.976          & 3.714          \\
Gemini-3.1-flash-lite \cite{gemini25}    & 0.242         & 0.406            & 0.225      &  0.059         &  0.072         & 0.049         & 0.174          & 3.198           & 3.737          & 3.492          \\
Qwen3.5-4B \cite{qwen3.5}                          & 0.309    & 0.231    & 0.276   & 0.051     & 0.074         & 0.040         & 0.142   & 2.699    & 3.491   & 3.037   \\
Qwen3.5-4B$_{\text{SFT}}$ (\textbf{Ours})            & \cellcolor{red!25} \textbf{0.897}    & \cellcolor{red!25} \textbf{0.576}    & \cellcolor{red!25} \textbf{0.354}   & \cellcolor{red!25} \textbf{0.190}     & \cellcolor{red!25} \textbf{0.482}         & \cellcolor{red!25} \textbf{0.429}         & \cellcolor{red!25} \textbf{0.451}   & \cellcolor{red!25} \textbf{3.741}    & \cellcolor{red!25} \textbf{4.238}   & \cellcolor{red!25} \textbf{3.746}  \\
\bottomrule
\end{tabular}
}
\end{table*}

\begin{table*}[t]
\centering
\footnotesize
\caption{Ablation study on Qwen2.5VL-7B for reward normalization and LLM judge. Row A: our full MedGRPO method with four task rewards. Rows B-E isolate individual contributions. Notation: task combinations (e.g., \texttt{TAG}+\texttt{STG}) indicate which task rewards are used during GRPO training.}
\label{tab:ablation}
\resizebox{\linewidth}{!}{
\begin{tabular}{@{}lcccccccccc@{}}
\toprule
Model            & \texttt{CVS}$_{acc}$ & \texttt{NAP}$_{acc}$ & \texttt{SA}$_{acc}$ & \texttt{STG}$_{mIoU}$ & \texttt{TAG}$_{mIoU@0.3}$ & \texttt{TAG}$_{mIoU@0.5}$ & $\texttt{DVC}_{F1}$ & \texttt{DVC}$_{llm}$ & \texttt{VS}$_{llm}$ & \texttt{RC}$_{llm}$ \\   \midrule
A \texttt{TAG}+\texttt{STG}+\texttt{VS}+\texttt{RC} (full)     &\cellcolor{orange!25}  0.896         &\cellcolor{orange!25} 0.405          
&\cellcolor{yellow!25} 0.254 
&\cellcolor{red!25}\bf 0.202               &\cellcolor{red!25} \bf 0.216         &\cellcolor{red!25} \bf 0.156         &\cellcolor{orange!25} 0.214          &\cellcolor{red!25} \bf 3.797           &\cellcolor{red!25} \bf 4.184          &\cellcolor{orange!25} 3.442          \\
\midrule
B \texttt{TAG}+\texttt{STG}+\texttt{VS}+\texttt{RC} w/o normal   & 0.020         & 0.267              & 0.234      & 0.010       & 0.004         & 0.003         & 0.002          & 1.061           &\cellcolor{yellow!25} 3.805          &  \cellcolor{red!25} \bf 3.469          \\
C \texttt{TAG}+\texttt{STG} w/o \texttt{VS}+\texttt{RC}   & \cellcolor{red!25} \bf 0.914         &\cellcolor{yellow!25} 0.394           & \cellcolor{orange!25}0.257           
&\cellcolor{orange!25} 0.193     
&\cellcolor{orange!25} 0.202         &\cellcolor{orange!25}  0.142         &\cellcolor{red!25} \bf 0.225          &\cellcolor{orange!25}  3.718           & 3.776          & \cellcolor{yellow!25} 3.425          \\
D \texttt{VS}+\texttt{RC} w/ llm-judge     &\cellcolor{yellow!25} 0.894         & \cellcolor{red!25}\bf 0.434           & 0.239 & \cellcolor{yellow!25} 0.183               &\cellcolor{yellow!25}  0.149         &\cellcolor{yellow!25} 0.096         & \cellcolor{yellow!25} 0.165          & \cellcolor{yellow!25} 3.688           &\cellcolor{orange!25}  3.824          & 3.235          \\
E \texttt{VS}+\texttt{RC} w/o llm-judge   & \cellcolor{yellow!25} 0.894 & 0.363  & \cellcolor{red!25} \bf 0.259 & \cellcolor{yellow!25} 0.183 & 0.140 & 0.090 & 0.161 & 3.628 & 3.733 & 2.984        \\
\bottomrule
\end{tabular}}
\vspace{-1em}
\end{table*}

\vspace{0.4em}
\noindent\textbf{LLM Judge Design.}
We design a GPT-4.1-based judge using \emph{comparative similarity scoring}: the judge evaluates ``How closely does the generated caption match the reference?'' rather than rating absolute quality. This comparative framing avoids score inflation~\cite{geval} and provides better discrimination between model qualities. The judge evaluates five clinical dimensions with explicit 1--5 rubrics~\cite{videochatgpt}: (1) \textit{medical terminology precision} (clinical terms vs. lay language), (2) \textit{instrument \& anatomy identification} (specific tools and structures), (3) \textit{specificity vs. vagueness} (precise details vs. generic descriptions), (4) \textit{clinical procedure context} (workflow and safety awareness), and (5) \textit{action \& state accuracy} (surgical actions and tissue states). For each dimension, scores range from 1 (``completely different'') to 5 (``semantically equivalent''). We compute the mean score $\bar{s} = \frac{1}{5}\sum_{i=1}^5 s_i$ and apply logistic normalization: $r_{\text{LLM}} = r_{\text{norm}}(\bar{s})$.

\vspace{0.4em}
\noindent\textbf{Hybrid Evaluation Design.}
To capture both overall semantic coherence and fine-grained medical correctness, we combine semantic similarity and LLM judge evaluation. For caption generation tasks, we compute the final reward as an equal-weighted average of normalized semantic similarity and LLM judge scores. This design provides three advantages: (1) \textit{complementary evaluation levels}---semantic similarity captures paragraph-level coherence while LLM judge assesses detail-level clinical correctness, (2) \textit{medical accuracy}---domain-specific evaluation captures clinical nuances invisible to semantic metrics, (3) \textit{robustness}---the 50\% weight on semantic similarity ensures meaningful reward signals even when LLM evaluation occasionally fails.

\section{Experiments}
\label{sec:experiments}

We conduct comprehensive experiments to evaluate MedGRPO across medical video understanding tasks. We demonstrate our approach achieves strong performance improvements over SFT baselines, enables dataset-fair optimization through reward normalization, and produces domain-specific improvements via LLM judge evaluation.

\subsection{Experimental Setup}
\label{sec:setup}

\vspace{0.4em}
\noindent\textbf{Datasets.} MedVidBench (detailed in §\ref{sec:dataset}) provides two versions designed for different purposes. The \textbf{Large-Scale version} contains 531,850 samples (461,413 train and 70,437 test across 626 videos) leveraging all 8 medical datasets, maximizing available training data but with natural task imbalance favoring captioning tasks. This version is valuable for scaling law experiments and scenarios where maximum data utilization is prioritized. The \textbf{Standard version} contains 51,505 samples (45,260 train and 6,245 test across 611 videos, representing 9.81\% of the large-scale version), carefully balanced across all tasks for efficient multi-task learning. Unless otherwise specified, all experiments use the Standard version for balanced evaluation.

\vspace{0.4em}
\noindent\textbf{Metrics.} We evaluate across three temporal granularities. \textbf{Video-level tasks} include Video Summarization (\texttt{VS}) measured by LLM judge score; Critical View of Safety (\texttt{CVS}) assessed by accuracy; Next Action Prediction (\texttt{NA}) evaluated by accuracy; and Skill Assessment (\texttt{SA}) measured by accuracy. \textbf{Segment-level tasks} include Temporal Action Grounding (\texttt{TAG}) measured by mean IoU at thresholds 0.3 and 0.5; Dense Video Captioning (\texttt{DVC}) evaluated by LLM judge score and F1 score; and Region Captioning (\texttt{RC}) assessed by LLM judge score. \textbf{Frame-level tasks} include Spatiotemporal Grounding (\texttt{STG}) measured by mIoU.

\vspace{0.4em}
\noindent\textbf{Baselines.} We compare our method against several baselines. \textbf{Off-the-shelf baselines} (GPT-4.1, Gemini-2.5-Flash, and Qwen2.5VL-7B) are evaluated using one-shot prompting with a format example but without any fine-tuning, performing significantly worse than fine-tuned models and demonstrating the difficulty of medical video understanding without domain adaptation. \textbf{Qwen2.5VL-7B$_\text{SFT}$} serves as our primary baseline, trained via supervised fine-tuning on MedVidBench. \textbf{Qwen2.5VL-7B$_\text{MedGRPO}$} represents our full method, which applies MedGRPO training to Qwen2.5VL-7B$_\text{SFT}$ with cross-dataset reward normalization and medical LLM judge design. We further validate generalizability by applying the same SFT and MedGRPO pipeline to \textbf{Qwen3-VL-4B}~\cite{qwen3vl}, a smaller model with improved temporal modeling. We also compare against 2026 off-the-shelf models (\textbf{GPT-5.4}, \textbf{Gemini-3.1-Flash-Lite}), evaluated with the same one-shot prompting, and apply SFT to \textbf{Qwen3.5-4B}~\cite{qwen3vl} to evaluate performance with the latest foundation models.

\begin{figure}[t]
\centering
\includegraphics[width=0.9\linewidth,height=3.5cm]{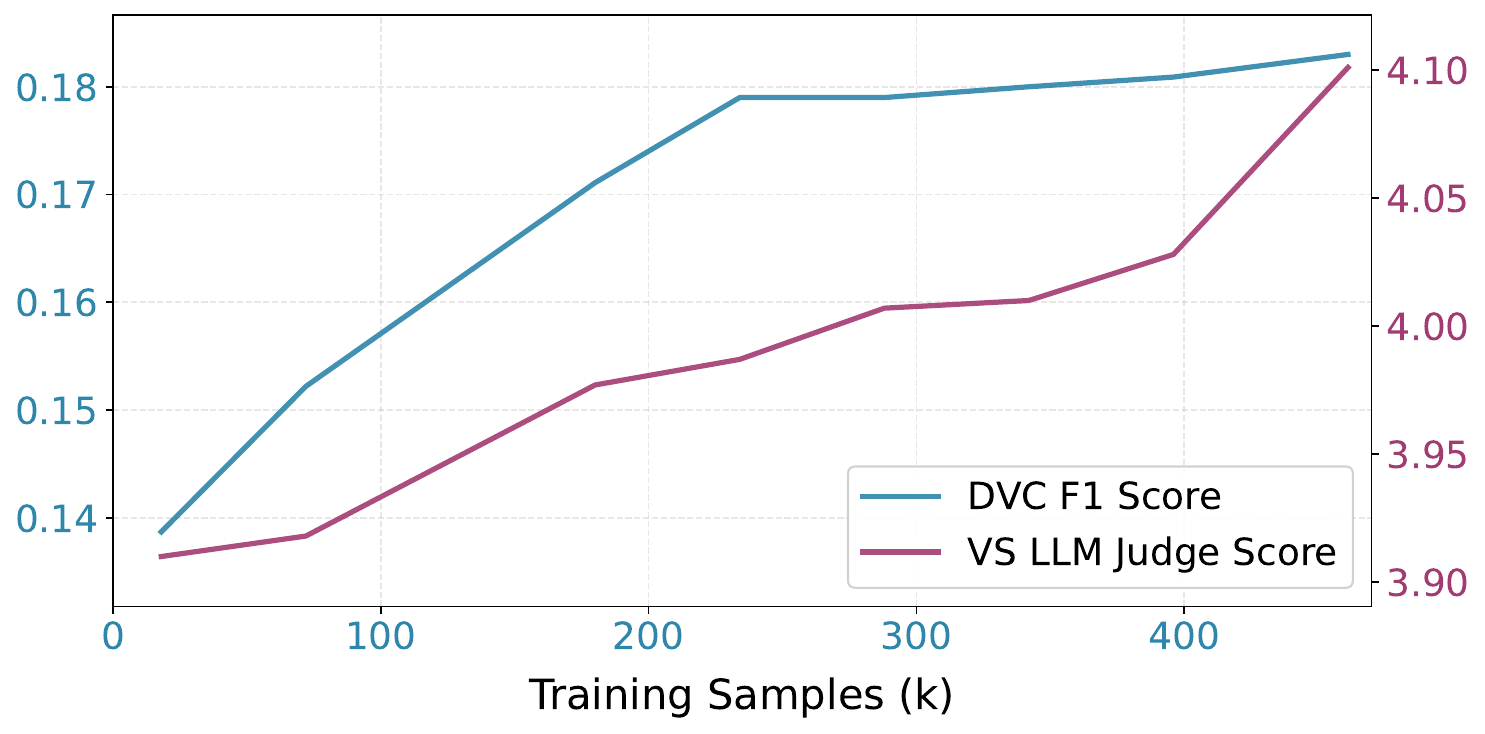}
\caption{Scaling law analysis. Performance on Dense Video Captioning (\texttt{DVC} F1 score) and Video Summarization (\texttt{VS} LLM judge score) improves consistently as training samples from the Large-Scale version increase from 0 to 461K.}
\label{fig:scale_law}
\vspace{-1.5em}
\end{figure}

\noindent\textbf{Implementation Details.} We use Qwen2.5-VL-7B as the primary base model. During SFT training, video frames are sampled at 0.1-3 FPS with adaptive sampling for longer procedures. During GRPO training, frames are sampled at 1 FPS. Each frame contains 8$\times$28$\times$28 to 48$\times$28$\times$28 pixels. SFT training on the Standard version runs for 3 epochs with learning rate 5e-7 and batch size 8, while scaling law experiments on the Large-Scale version use 1 epoch to demonstrate performance improvements as training data increases from 0 to 461K samples (Figure~\ref{fig:scale_law}). GRPO training uses 5000 gradient updates with learning rate 5e-7 and batch size 5. We adopt technical improvements from DAPO~\cite{dapo} in our GRPO implementation with group size $G=8$ and clipping parameters $\epsilon_\text{low}=0.2$ and $\epsilon_\text{high}=0.28$. Logistic normalization uses slope $k=3.0$ and format penalty $\beta=0.6$. All experiments run on 8$\times$ H100 GPUs.

\begin{figure*}[t]
\centering
\includegraphics[width=\linewidth,height=9cm]{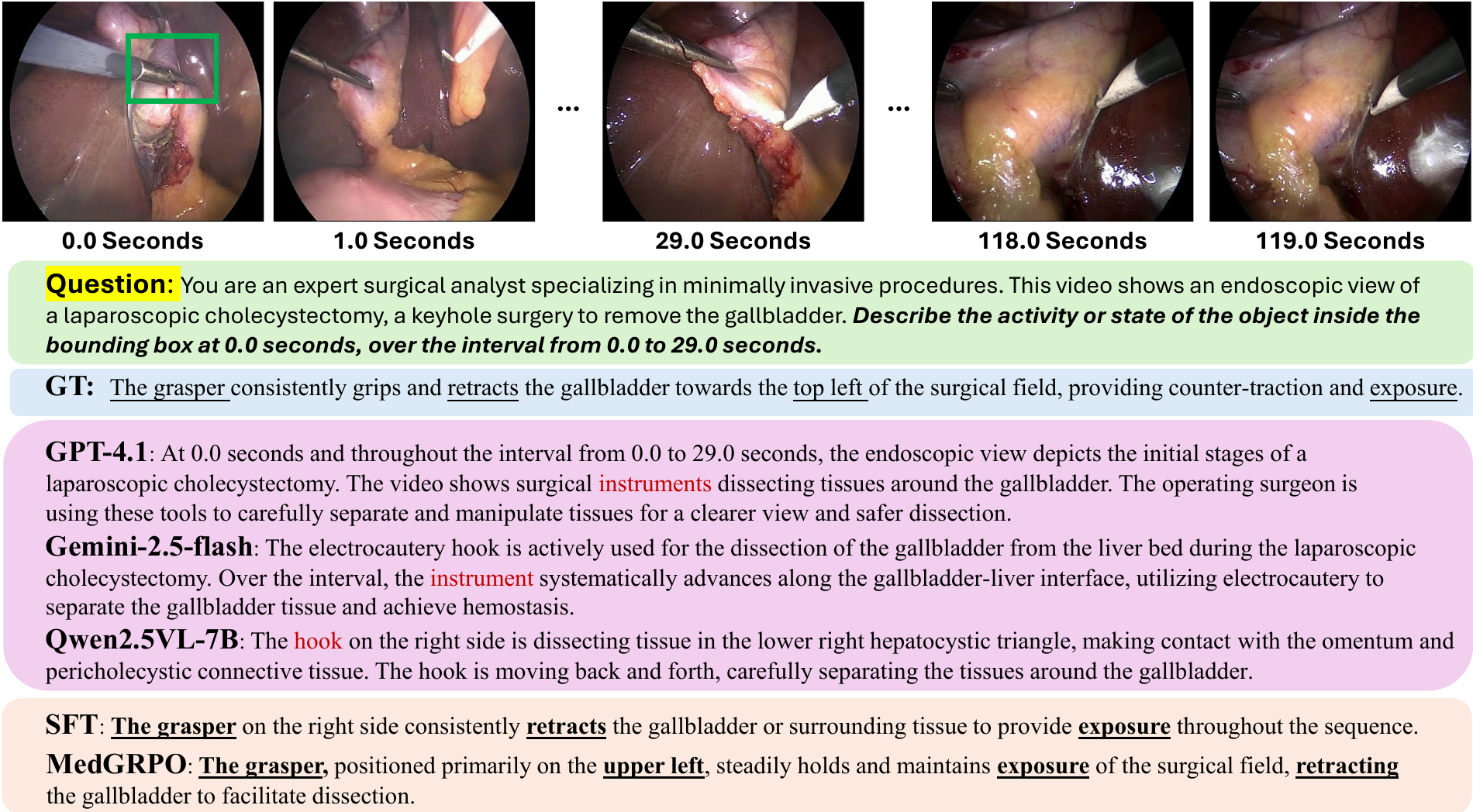}
\caption{Qualitative comparison of region captioning generation.}
\label{fig:qualitative}
\vspace{-1em}
\end{figure*}

\subsection{Main Results}
\label{sec:main_results}
Table~\ref{tab:main_results} presents our main results across all 8 tasks. We compare against off-the-shelf baselines from 2025 (GPT-4.1, Gemini-2.5-Flash, VideoChat-R1.5-7B, Qwen2.5VL-7B, Qwen2.5VL-7B$_{\text{Surg-CholecT50}}$, and Qwen3-VL-4B) and 2026 (GPT-5.4, Gemini-3.1-Flash-Lite), our SFT baselines, and our full MedGRPO method on Qwen2.5VL-7B, Qwen3-VL-4B, and Qwen3.5-4B.
\vspace{0.4em}

\noindent\textbf{Comparison with Off-the-shelf Baselines.}
Our SFT baseline significantly outperforms all off-the-shelf baselines, demonstrating the importance of domain adaptation.
GPT-4.1 achieves only 0.018 on \texttt{CVS} and 0.014 on \texttt{STG}, while Gemini-2.5-Flash reaches 0.101 on \texttt{CVS} and 0.047 on \texttt{STG}.
VideoChat-R1.5-7B, a recent RL-trained general-domain VideoLLM, scores 0.000 on \texttt{CVS} and \texttt{STG}, showing that general video RL training does not transfer to medical tasks.
Off-the-shelf Qwen2.5VL-7B similarly performs poorly across grounding tasks (\texttt{STG}: 0.020, \texttt{TAG}@0.3: 0.006).
Qwen2.5VL-7B$_{\text{Surg-CholecT50}}$, a recent medical VideoLLM fine-tuned on laparoscopic data, also fails to generalize (\texttt{CVS}: 0.000, \texttt{STG}: 0.000), highlighting that single-dataset surgical specialization is insufficient for the diverse tasks in MedVidBench.
In contrast, our SFT baseline achieves 0.894 on \texttt{CVS} and 0.177 on \texttt{STG}, establishing the importance of large-scale multi-source training on MedVidBench.

\vspace{0.4em}
\noindent\textbf{MedGRPO Performance.} Our full method (Qwen2.5VL-7B$_{\text{MedGRPO}}$) achieves consistent improvements over SFT across most tasks: \texttt{CVS} (+0.002 to 0.896), \texttt{STG} (+0.025 to 0.202), \texttt{SA} (+0.036 to 0.254), \texttt{TAG}@0.3 (+0.074 to 0.216), \texttt{TAG}@0.5 (+0.065 to 0.156), \texttt{DVC}$_{\text{llm}}$ (+0.132 to 3.797), \texttt{DVC}$_\text{F1}$ (+0.049 to 0.214), \texttt{VS}$_{\text{llm}}$ (+0.588 to 4.184), and \texttt{RC}$_{\text{llm}}$ (+0.685 to 3.442). \texttt{NAP} slightly decreases from 0.442 to 0.405, as it was not included in the reward optimization (we use \texttt{TAG}, \texttt{STG}, \texttt{VS}, and \texttt{RC} for computing rewards). Caption generation tasks benefit most from our hybrid LLM judge design, with \texttt{VS} and \texttt{RC} showing substantial improvements. Temporal grounding tasks also show strong gains, with \texttt{TAG}@0.3 and \texttt{TAG}@0.5 improving significantly.

\vspace{0.4em}
\noindent\textbf{Generalization to Qwen3-VL-4B.} To validate framework generalizability, we apply the same SFT and MedGRPO pipeline to Qwen3-VL-4B~\cite{qwen3vl}, a smaller model with improved temporal modeling. Off-the-shelf Qwen3-VL-4B achieves near-zero performance on grounding tasks (\texttt{STG}: 0.000, \texttt{CVS}: 0.000), confirming that architectural advances alone cannot address medical video understanding without domain adaptation. Our SFT training yields strong gains (\texttt{CVS}: 0.895, \texttt{TAG}@0.3: 0.465, \texttt{TAG}@0.5: 0.403), and MedGRPO further improves across all tasks, with notable gains on \texttt{STG} (+0.043), \texttt{TAG}@0.3 (+0.039), and \texttt{DVC}$_\text{F1}$ (+0.045). These results demonstrate that MedGRPO generalizes across model architectures and scales, consistently improving upon SFT baselines regardless of the underlying model.

\vspace{0.4em}
\noindent\textbf{Comparison with 2026 Models.} We further compare against the latest 2026 off-the-shelf models: GPT-5.4, Gemini-3.1-Flash-Lite, and Qwen3.5-4B. While these models show substantial improvements over their 2025 counterparts (e.g., GPT-5.4 reaches 0.164 on \texttt{CVS} vs. 0.018 for GPT-4.1, and 3.976 on \texttt{VS} vs. 2.490), they still fall significantly short of our fine-tuned models on grounding tasks (\texttt{STG}: 0.004, 0.059, and 0.051, \texttt{TAG}@0.3: 0.086, 0.072, and 0.074). Notably, off-the-shelf Qwen3.5-4B achieves 0.309 on \texttt{CVS} and 3.491 on \texttt{VS}, substantially outperforming its 2025 predecessor Qwen3-VL-4B (0.000 and 2.926), yet still far below our SFT models. Applying SFT to Qwen3.5-4B achieves the highest \texttt{NAP} accuracy (0.576) across all models and strong performance on grounding (\texttt{TAG}@0.3: 0.482, \texttt{TAG}@0.5: 0.429), further confirming that domain adaptation on MedVidBench remains essential even as foundation models advance. MedGRPO training on Qwen3.5-4B will be added in a future update.

\vspace{0.4em}
\noindent\textbf{Qualitative Analysis.} Figure~\ref{fig:qualitative} illustrates representative improvements in region caption generation. Ground Truth describes: ``The grasper consistently grips and retracts the gallbladder towards the top left of the surgical field, providing counter-traction and exposure." GPT-4.1 produces generic descriptions without specific instruments. Gemini-2.5-Flash misidentifies the tool as ``electrocautery hook" with incorrect actions. Our SFT baseline identifies ``grasper" but uses vague spatial terms (``right side"). MedGRPO generates clinically accurate captions: ``The grasper, positioned primarily on the upper left, steadily holds and maintains exposure of the surgical field, retracting the gallbladder to facilitate dissection," demonstrating precise instrument identification, accurate spatial localization, specific action descriptions, and clinical context understanding.

\subsection{Ablation Studies}
\label{sec:ablations}

We conduct ablation studies on Qwen2.5VL-7B to validate each component.

\vspace{0.4em}
\noindent\textbf{Reward Normalization.}
Table~\ref{tab:ablation} Row B shows that the removal of reward normalization causes catastrophic training collapse: performance drops from 0.894 (SFT baseline) to 0.020 in \texttt{CVS}, 0.177 to 0.010 in \texttt{STG} and 0.142 to 0.004 in \texttt{TAG}@0.3. Figure~\ref{fig:method_overview}(b) visualizes this collapse through training entropy, showing that without reward normalization, entropy becomes highly unstable with dramatic spikes, while our reward normalization maintains stable entropy throughout training. This occurs because unnormalized rewards create high magnitude differences between dataset-task pairs, causing the optimizer to focus exclusively on high magnitude tasks while neglecting others. In contrast, Row A shows that our full method with reward normalization maintains balanced improvements across all tasks, confirming that proper reward scaling is essential for stable multi-dataset RL training.

\vspace{0.5em}
\noindent\textbf{Multi-Task Learning Benefits.}
Comparing Row A (full method with all four types of task) versus Row C (only \texttt{TAG}+\texttt{STG} without \texttt{VS}+\texttt{RC}), we observe that training with caption generation tasks improves grounding performance: \texttt{STG} improves from 0.193 to 0.202 (+4.7\%), \texttt{TAG}@0.3 from 0.202 to 0.216 (+6.9\%), and \texttt{TAG}@0.5 from 0.142 to 0.156 (+9.9\%). This demonstrates beneficial task synergy in multi-task learning: caption generation requires understanding video content and temporal dynamics, which creates richer visual representations that benefit spatial-temporal localization tasks. Similarly, grounding tasks help captioning by providing better spatial awareness.

\vspace{0.5em}
\noindent\textbf{LLM Judge Design.}
Rows D-E isolate the contribution of our medical LLM judge for caption generation tasks. Row D (\texttt{VS}+\texttt{RC} with LLM judge) improves \texttt{VS} from 3.596 (SFT baseline) to 3.824 and \texttt{RC} from 2.757 to 3.235, demonstrating that domain-specific evaluation captures medical precision beyond semantic similarity. Row E (\texttt{VS}+\texttt{RC} without LLM judge) achieves only 3.733 on \texttt{VS} and 2.984 on \texttt{RC}, showing the importance of medical-specific evaluation. The full method (Row A) combining all four task rewards achieves the strongest performance: \texttt{VS} 4.184 and \texttt{RC} 3.442, improving 16.4\% and 24.8\% over SFT baseline. 
\section{Conclusion}
\label{sec:conclusion}
\vspace{-0.2em}
We present MedVidBench, a large-scale heterogenous benchmark for medical video understanding, and MedGRPO, a multi-dataset RL framework. We find that standard RL collapses on this data due to unbalanced reward scales across datasets. MedGRPO prevents this using two innovations: cross-dataset reward normalization to balance optimization, and a medical LLM judge for domain-specific evaluation. Our method substantially outperforms supervised fine-tuning, establishing that multi-dataset RL requires reward balancing and domain-adapted evaluation.
\vspace{-1em}
{
    \small
    \bibliographystyle{ieeenat_fullname}
    \bibliography{main}
}
\appendix
\twocolumn[\centering \section*{\Large Supplementary Material: \\ MedGRPO: Multi-Task Reinforcement Learning for \\ Heterogeneous Medical Video Understanding }]

This supplementary material provides comprehensive details on MedVidBench dataset construction and MedGRPO training methodology. \S\ref{sec:supp_curation} describes our data curation pipeline, including prompt design strategies for web-sourced and frame-annotated datasets, QA generation procedures, and human validation study results. \S\ref{sec:supp_statistics} presents detailed dataset statistics covering task distribution, temporal characteristics, and annotation quality patterns. \S\ref{sec:supp_implementation} provides implementation details for SFT, GRPO training, and skill assessment evaluation. \S\ref{sec:supp_llm_judge} details our medical LLM judge rubrics across five clinical evaluation dimensions. \S\ref{sec:supp_qualitative} presents additional qualitative results and failure analysis. All materials support reproducibility and provide insights beyond the main paper's scope constraints.

\section{Dataset Curation Pipeline}
\label{sec:supp_curation}

Our data curation pipeline transforms existing expert annotations from 8 medical video datasets into high-quality instruction-following format using dual multi-modal large language models (GPT-4.1~\cite{gpt4v}  and Gemini-2.5-Flash~\cite{gemini25}). We employ a multi-perspective approach adapting prompting strategies to dataset characteristics: frame-annoated datasets (CholecT50~\cite{cholect50}, EgoSurgery~\cite{egosurgery}, CholecTrack20~\cite{cholectrack20}, CoPESD~\cite{copesd}) receive rich contextual prompts incorporating frame-level annotations, while web-sourced datasets (AVOS~\cite{avos}, NurViD~\cite{nurvid}) utilize high-quality audio transcripts (Whisper-X~\cite{whisperx}) and video metadata to supplement visual understanding.

\subsection{Prompt Design Principles}

Our prompts follow a consistent six-component structure: (1) \textbf{role definition} establishing domain expertise (Expert medical analyst), (2) \textbf{background knowledge} providing procedure-specific context (anatomy, key structures, workflow), (3) \textbf{input data specification} enumerating available information (frames, timestamps, annotations), (4) \textbf{task definition} clarifying the objective (generate temporal summary, describe region), (5) \textbf{guiding principles} enforcing quality standards (be visually grounded, use precise terminology, avoid verbatim copying), and (6) \textbf{output format} specifying structure (one sentence, emphasize dynamics). We instantiate this template differently based on available annotations, as shown below.

\subsection{Web-Sourced Datasets}

For Web-sourced datasets, we compensate for limited expert annotations through multi-modal context integration. We enrich prompts with video metadata (title) and segment level annotation (action labels), high-quality temporal aligned ASR transcripts extracted using Whisper-X to provide comprehensive context. The prompt template is as below:

\begin{tcolorbox}[colback=gray!5,colframe=gray!40,boxrule=0.5pt,arc=2pt,left=3pt,right=3pt,top=3pt,bottom=3pt]
\small
\textbf{Role:} Expert video analyst specializing in medical procedures

\medskip
\noindent\textbf{Background Knowledge:}
\begin{itemize}[leftmargin=*,noitemsep,topsep=0pt]
\item Video Title
\item Video Description
\end{itemize}

\medskip
\noindent\textbf{Input Data:}
\begin{itemize}[leftmargin=*,noitemsep,topsep=0pt]
\item Frame\_i -- raw video frame
\item Timestamp\_i -- normalized value in [0, 1]
\item Action Label
\item Transcript Segments: [timestamp interval: text]
\item Context: action label
\end{itemize}

\medskip
\noindent\textbf{Task:} Analyze frames and generate concise summary describing temporal evolution

\medskip
\noindent\textbf{Guiding Principles:}
\begin{itemize}[leftmargin=*,noitemsep,topsep=0pt]
\item Be Visually Grounded: Focus on observable events only
\item Use Precise Naming: Specific surgical terminology
\item Avoid Verbatim Copying: No prompt phrases in output
\item Be Concise and Direct: No generic filler
\item Focus on Dynamics: Object movement and instrument actions
\end{itemize}

\medskip
\noindent\textbf{Output Format:} One sentence describing what happens over time, emphasizing motion, interaction, and anatomical changes.
\end{tcolorbox}

\subsection{Frame-Annotated Datasets}

For datasets with rich expert annotations, we maximize information utilization through two complementary annotation strategies: (1) frame-text interleaved input. Frame-wise texts including triplet annotations (CholecT50) providing surgical action labels as (instrument, verb, target) triplets and textual descriptions (CoPESD) providing detailed per-frame narrative annotations. (2) bounding box visual prompts (CholecTrack20, EgoSurgery) overlaying spatial object locations with object labels directly on frames. The prompt template is as below:

\begin{tcolorbox}[colback=gray!5,colframe=gray!40,boxrule=0.5pt,arc=2pt,left=3pt,right=3pt,top=3pt,bottom=3pt,breakable]
\small
\textbf{Role:} Expert surgical analyst

\medskip
\noindent\textbf{Background Knowledge}: Per-surgery background info for each dataset, e.g., Laparoscopic Cholecystectomy
\begin{itemize}[leftmargin=*,noitemsep,topsep=0pt]
\item Anatomy: Gallbladder anatomy and position
\item Key Structures: Cystic duct, common bile duct, cystic artery
\item Critical Landmark: Hepatocystic triangle (Calot triangle)
\end{itemize}

\medskip
\noindent\textbf{Input Data:}
\begin{itemize}[leftmargin=*,noitemsep,topsep=0pt]
\item (Interleaved) Frame\_i -- video frame with bounding box overlay and per-frame annotation
\item Timestamp\_i -- normalized [0, 1]
\item Action Label
\end{itemize}

\medskip
\noindent\textbf{Task:} Analyze sequence and generate concise summary describing temporal evolution

\medskip
\noindent\textbf{Guiding Principles:}
\begin{itemize}[leftmargin=*,noitemsep,topsep=0pt]
\item Be Visually Grounded: Observable events only
\item Use Precise Naming: Specific instrument names from annotations
\item Use Preferred Verbs: Verbs from annotation vocabulary
\item Be Concise and Direct: No filler
\item Focus on Dynamics: Object movement, instrument causation
\end{itemize}

\medskip
\noindent\textbf{Output Format:} One sentence describing what visibly happens over time
\end{tcolorbox}

For regional captioning, we adapt the same prompt template by modifying the task specification to generate per-object descriptions with emphasis on spatial location and object-specific movements.

\subsection{QA Generation and Quality Assurance}

After caption generation, we create diverse QA pairs by combining dataset-specific context prefixes with task-specific question templates (3--6 template variants per task), producing instruction-following instances for all 8 tasks spanning video-level, segment-level, and frame-level understanding. To ensure annotation quality, we employ dual-model validation: for caption generation tasks (video summary, dense captioning, region captioning), we independently generate captions using both GPT-4.1 and Gemini-2.5-Flash, compute semantic similarity using sentence-transformers, and filter low-quality pairs with similarity $<$0.3. This dual-model approach prevents model-specific biases and hallucinations while ensuring consistent high-quality annotations. For evaluating caption quality during both dataset validation and RL training, we design a medical LLM judge (detailed rubrics in \S\ref{sec:supp_llm_judge}) that assesses captions across five clinical dimensions through comparative similarity scoring.

\subsection{Human Validation Study}

\begin{figure*}[t]
\centering
\includegraphics[width=\linewidth]{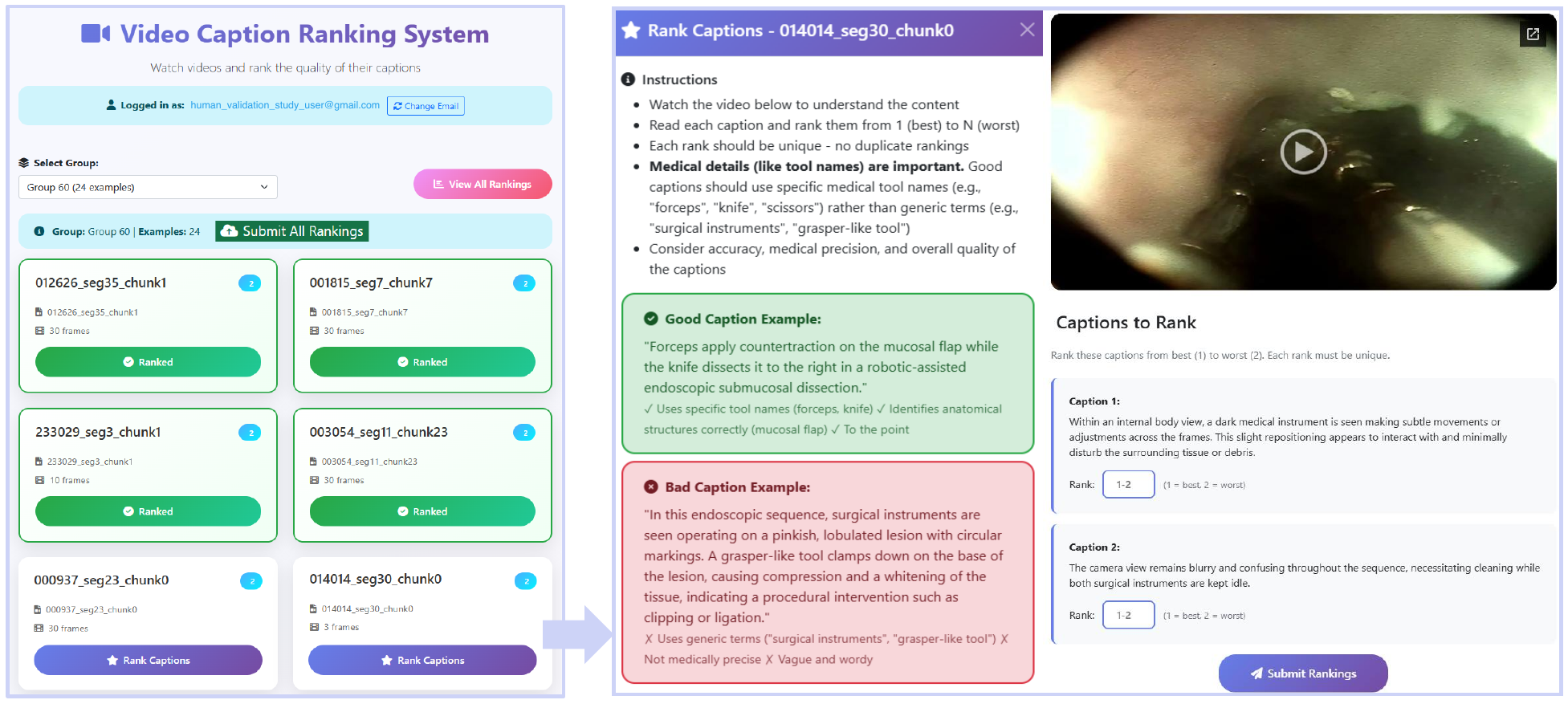}
\caption{{Interface for human validation study.} Users were provided detailed instruction to rank caption after watching a short video. An instruction example for a good and bad caption was provided. }
\label{fig:userstudy_website}
\end{figure*}

\begin{figure}[th]
\centering
\includegraphics[width=\linewidth]{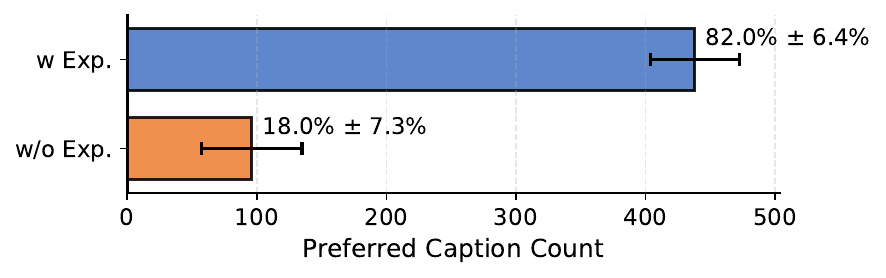}
\caption{{Human validation study results.} User preference comparison with 12 participants on CoPESD dataset. ``w/ Expert Prompt'' refers to captions generated using our annotation-enriched prompting with overlaid bounding boxes, procedure context, and expert annotations. ``w/o Expert Prompt'' refers to captions generated from raw frames only with minimal prompting. Participants strongly prefer captions generated with expert prompts (82.0\% vs 18.0\%), validating our multi-perspective quality assurance pipeline.}
\label{fig:userstudy}
\end{figure}

To validate our annotation-enriched prompting approach, we conducted a user study with 12 participants  who are experts, work in medical data analysis, to compare captions generated using our expert prompts versus a frames-only baseline, both of which are described next. For CoPESD~\cite{copesd} dataset, we generated two types of captions: (1) \textbf{with expert prompt}: captions using our full pipeline with overlaid bounding boxes, procedure-specific context, and expert annotations; (2) \textbf{without expert prompt}: captions generated from raw video frames only using a minimal prompt (``Describe what you see in this healthcare procedure video in one sentence''), without procedural context, annotation overlays, timestamps, or domain knowledge. We developed a web interface, shown in Figure~\ref{fig:userstudy_website}, to allow participants to rank the caption-pairs. Participants were provided with detailed instructions and examples to select the superior caption based on clinical accuracy and terminological precision. Figure~\ref{fig:userstudy} shows the results: participants strongly preferred captions generated with expert prompts (82.0\%) over frames-only captions (18.0\%), confirming that our annotation-enriched prompting strategy produces superior medical video descriptions compared to naive frame-based generation.

\section{MedVidBench: Dataset Statistics}
\label{sec:supp_statistics}

\paragraph{Task and Domain Distribution.}
Table~\ref{tab:dataset_stats} shows the statistical breakdown of MedVidBench by task and dataset. MedVidBench covers 8 dataset sources and 8 tasks spanning 4 domains: laparoscopic surgery (184.5K samples, 34.7\%), open surgery (216.8K, 40.8\%), robotic surgery (1.0K, 0.2\%), and nursing (129.5K, 24.4\%). The task distribution in MedVidBench reflects annotation granularity: frame-level annotations (e.g., spatial boxes) enable abundant region captioning samples (210.3K, 39.5\%) as each frame contains multiple annotated regions, while specialized tasks at video level requiring expert holistic assessment remain rare (skill assessment: 1.0K, 0.2\%, CVS: 4.4k, 0.8\%). Segment-level tasks like temporal action grounding (158.5K, 29.8\%) and dense captioning (73.3K, 13.8\%) fall between these extremes.

\begin{figure*}[t]
\centering
\includegraphics[width=\linewidth]{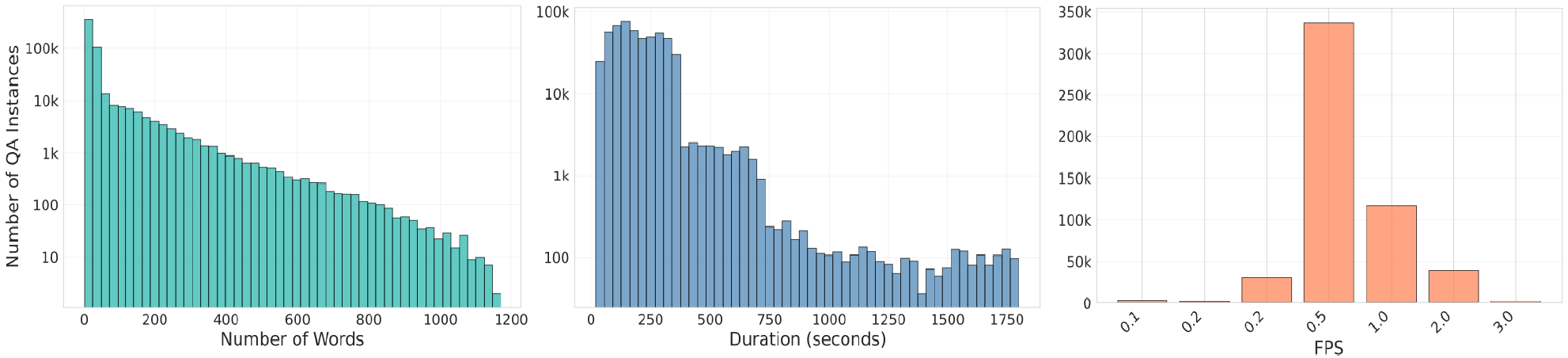}
\caption{{Dataset distribution analysis.} Dataset distribution across 532K QA instances from 8 medical video datasets. (Left) Answer length distribution showing word counts ranging from 1 to 1,170 words (median: 21, mean: 41). Short answers ($\le$5 words, 28.1\%) are predominantly from temporal action grounding tasks, while long answers ($>$20 words, 51.8\%) come mainly from dense video captioning and region captioning tasks. (Middle) Video duration distribution showing durations from 20 to 1,800 seconds (median: 182s, mean: 212s), exhibiting a long-tail pattern. (Right) FPS distribution showing that most instances use 0.5 FPS (63.3\%), followed by 1.0 FPS (22.0\%) and 2.0 FPS (7.5\%). Left and middle
panels use logarithmic scale on y-axis; right panel uses linear scale.}
\label{fig:dataset_distribution}
\end{figure*}

\paragraph{Temporal Characteristics and Frame Sampling.}
Figure \ref{fig:dataset_distribution} (middle and right) shows MedVidBench exhibits substantial temporal diversity. Video durations range from 20 seconds to 1,800 seconds (30 minutes) with a median of 182 seconds and mean of 212 seconds, displaying a long-tail distribution where most videos fall within typical medical procedure segment lengths. Frame sampling rates vary from 0.1 to 3.0 FPS, with the majority of instances (63.3\%) using 0.5 FPS, followed by 1.0 FPS (22.0\%) and 2.0 FPS (7.5\%). This distribution reflects two key factors: (1) source datasets have varying native frame rates, and (2) task-specific temporal requirements differ substantially. Video-level tasks (e.g. video summary) analyze longer durations and thus use low sampling rates to maintain manageable frame sequence lengths while capturing procedural evolution. This adaptive sampling strategy accommodates both dataset constraints and task-specific temporal granularity requirements.

\begin{table*}[ht]
\centering
\caption{{MedVidBench statistics by dataset and task.} Our benchmark covers 8 medical video sources with 532K video-instruction pairs across 8 tasks spanning video-level, segment-level, and frame-level understanding. Task abbreviations: \texttt{VS} (Video Summarization), \texttt{SA} (Skill Assessment), \texttt{NAP} (Next Action Prediction), \texttt{CVS} (Critical View of Safety), \texttt{DVC} (Dense Video Captioning), \texttt{RC} (Region Captioning), \texttt{TAG} (Temporal Action Grounding), \texttt{STG} (Spatiotemporal Grounding).}
\resizebox{\linewidth}{!}{
\begin{tabular}{l|l|r|cccc|ccc|c|r}\toprule
\textbf{} &\textbf{} &\textbf{} &\multicolumn{4}{c|}{\textbf{Video-Level}} &\multicolumn{3}{c|}{\textbf{Segment-Level}} &\textbf{Frame-Level} &\textbf{} \\
\cline{4-11}
\textbf{Dataset} &\textbf{Domain} &\textbf{Videos} &\texttt{VS} &\texttt{SA} &\texttt{NAP} &\texttt{CVS} &\texttt{DVC} &\texttt{RC} &\texttt{TAG} &\texttt{STG} &\textbf{Total} \\
\midrule
\textbf{CholecT50}~\cite{cholect50} &Laparoscopic & 50 &\checkmark & - & \checkmark & - & \checkmark & - & \checkmark & - & 7.1K \\
\textbf{CholecTrack20}~\cite{cholectrack20} & Laparoscopic & 20 & - & - & - & - & - & \checkmark & - & \checkmark & 102.7K \\
\textbf{Cholec80-CVS}~\cite{cholec80cvs} & Laparoscopic & 80 & - & - & - & \checkmark & - & - & - & - & 4.4K \\
\textbf{CoPESD}~\cite{copesd} &Laparoscopic & 40 &\checkmark & - & \checkmark & - & \checkmark & \checkmark & \checkmark & \checkmark & 70.3K \\
\textbf{AVOS}~\cite{avos} &Open Surgery & 25 & - & - & \checkmark & - & \checkmark & - & \checkmark & - & 62.5K \\
\textbf{EgoSurgery}~\cite{egosurgery} &Open Surgery & 21 & - & - & - & - & - & \checkmark & - & \checkmark & 154.3K \\
\textbf{JIGSAWS}~\cite{JIGSAWS} &Robotic Surgery & 103 & - & \checkmark & - & - & - & - & - & - & 1.0K \\
\textbf{NurViD}~\cite{nurvid} &Nursing & 287 &\checkmark & - & \checkmark & - & \checkmark & - & \checkmark & - & 129.5K \\
\midrule
\textbf{Total samples} & & \textbf{626} & \textbf{6.8K} & \textbf{1.0K} & \textbf{9.5K} & \textbf{4.4K} & \textbf{73.3K} & \textbf{210.3K} & \textbf{158.5K} & \textbf{68.0K} & \textbf{531.8K} \\
\bottomrule
\end{tabular}}
\label{tab:dataset_stats}
\end{table*}

\begin{figure}
    \centering
    \includegraphics[width=\linewidth, trim={0cm 0cm 0.5cm 0cm}, clip]{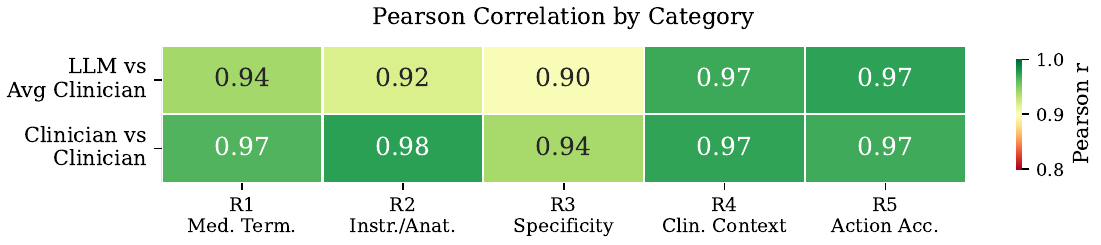}
    \caption{Pearson's correlation between average clinician-clinician ratings and LLM-Clinician ratings across five evaluation dimensions (\S~\ref{sec:supp_llm_judge:quality}). Experiment was performed with 10 board certified clinicians. The five dimensions are detailed in \S~\ref{sec:supp_llm_judge:rubrics}.}
    \label{fig:clinician_LLM}
\end{figure}

\begin{figure*}[t]
\centering
\includegraphics[width=\linewidth]{figures/task_vis.pdf}
\caption{{Examples of diverse tasks.} 5 diverse tasks from MedVidBench (Dense Video Captioning, Spatio-Temporal Grounding, Critical View Safety, Video Summary, and Next Action Prediction) spanning 3 domains (Nursing, Laparoscopic Surgery and Open Surgery).}
\label{fig:more_task}
\vspace{-1em}
\end{figure*}

\begin{figure*}[t]
\centering
\includegraphics[width=\linewidth]{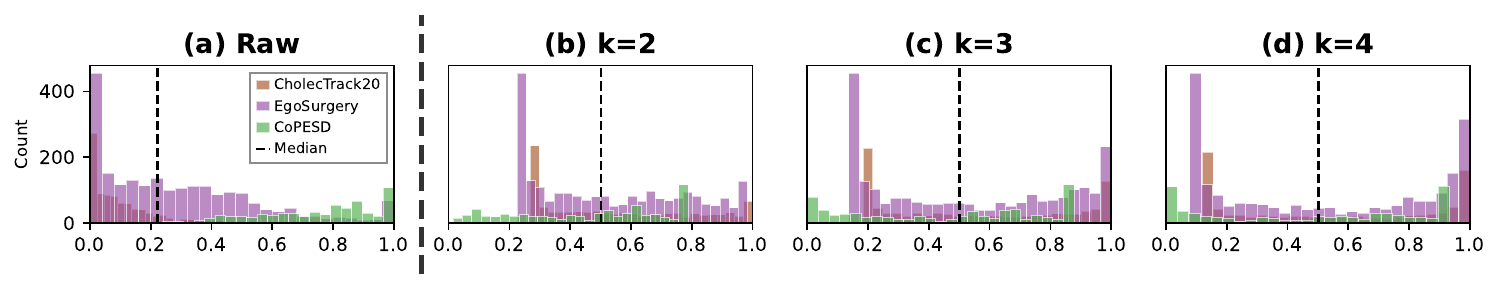}
\caption{Reward distributions of w/o and w/ normalization $k\in\{2, 3, 4\}$.}
\vspace{-1.em}
\label{fig:stg_k}
\end{figure*}

\paragraph{Annotation Quality and Word Counting.}
Figure \ref{fig:dataset_distribution} (left) shows the answer length distribution ranges from 1 to 1,170 words with a median of 21 words and mean of 41 words. Short answers ($\le$5 words, 28.1\%) are predominantly from temporal action grounding tasks providing concise timestamps. Long answers ($>$20 words, 51.8\%) come mainly from descriptive tasks, with dense video captioning generating the longest responses due to detailed narration of multiple sequential actions, followed by region captioning describing surgical instrument movements. This distribution reflects the fundamental task heterogeneity in medical video understanding: grounding tasks require precise localization with minimal text, while captioning tasks demand rich and accurate clinical descriptions.

\begin{figure*}[t]
\centering
\includegraphics[width=0.98\linewidth]{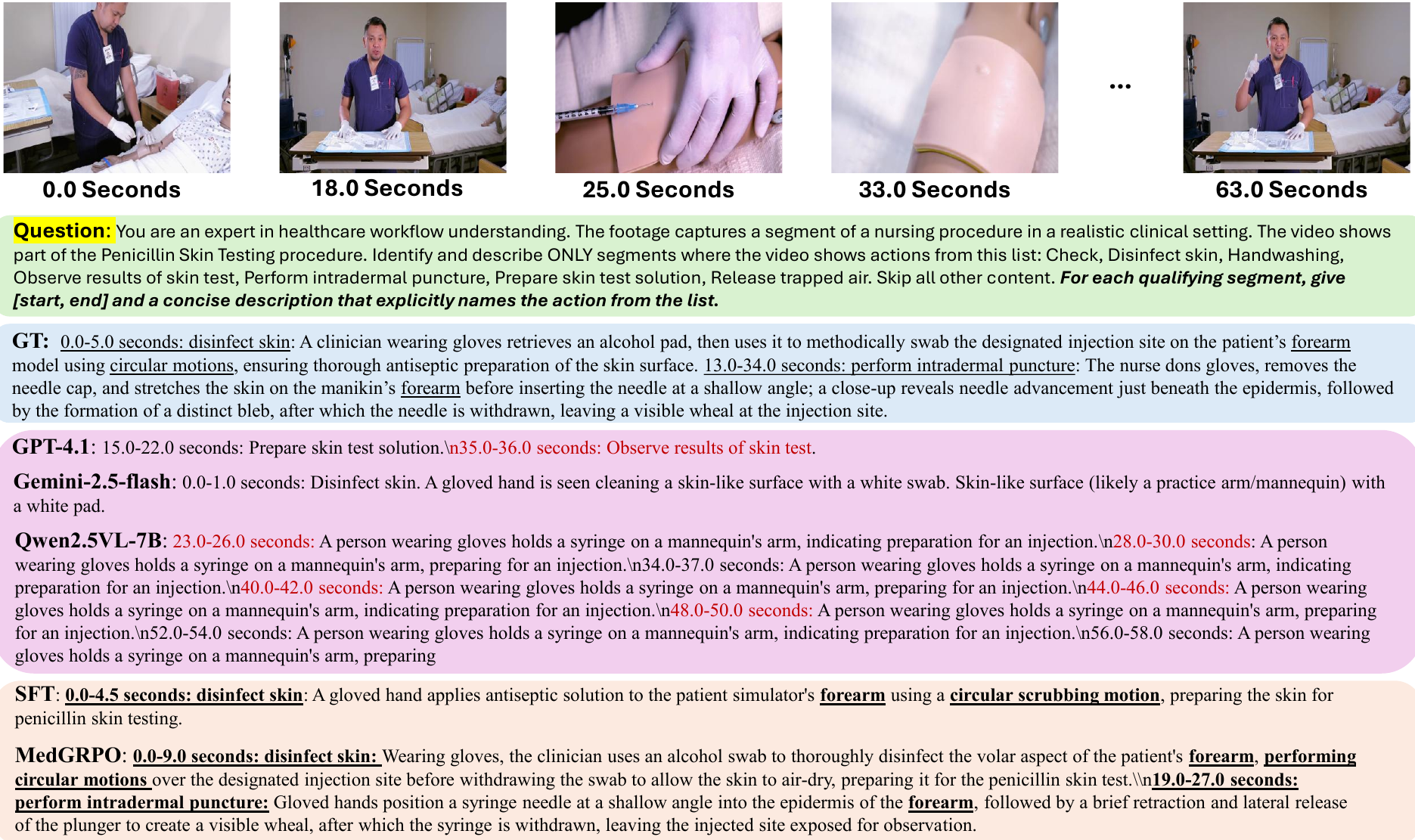}
\caption{Qualitative examples on dense video captioning.}
\label{fig:qualitative_2}
\vspace{-0.5em}
\end{figure*}

\section{Implementation Details}
\label{sec:supp_implementation}

\paragraph{Qwen2.5VL SFT Training.}
We use Qwen2.5-VL-7B-Instruct as our base model. Training is conducted on 8$\times$ H100 GPUs using distributed training with DeepSpeed ZeRO-3 offload. The per-device batch size is 6 with gradient accumulation steps of 1. We train for 3 epochs with differentiated learning rates: $5 \times 10^{-7}$ for the language model, $1 \times 10^{-6}$ for both the vision encoder and multimodal projector. A cosine learning rate scheduler is applied with a warmup ratio of 0.03. Weight decay is set to 0.01 and maximum gradient norm is clipped at 1.0. All training uses bfloat16 mixed precision. Video per-frame min and max pixels are set between $8 \times 28 \times 28$ to $48 \times 28 \times 28$ pixels. We fine-tune all model components including the vision encoder, multimodal projector, and language model, and enable gradient checkpointing to reduce memory usage.

\paragraph{Qwen2.5VL GRPO Training.}
We implement GRPO training using the EasyR1 framework built on veRL. Training is conducted on 8$\times$ H100 GPUs with the SFT checkpoint as initialization. We use a group size of $G=8$ responses per prompt with temperature 0.8 and top-p sampling at 0.95. The learning rate is $5 \times 10^{-7}$ and maximum gradient norm clipped at 0.5. Videos are sampled at 1.0 FPS to simplify exploration during rollout. Following DAPO practices, we use asymmetric PPO clipping with $\epsilon_\text{low}=0.2$ and $\epsilon_\text{high}=0.28$ and disable KL divergence penalty.

\paragraph{Skill Assessment Evaluation.}
We average the 6 OSATS dimension scores and apply thresholds to derive 3 classes (Novice/Intermediate/Expert). We report macro-averaged accuracy (mAcc) and MAE (lower is better): zero-shot Qwen2.5VL-7B achieves MAE=2.440, mAcc=0.000; SFT improves to MAE=1.262, mAcc=0.197; MedGRPO further improves to MAE=1.246, mAcc=0.254, demonstrating the effectiveness of our method.

\begin{figure*}[t]
\centering
\includegraphics[width=0.98\linewidth]{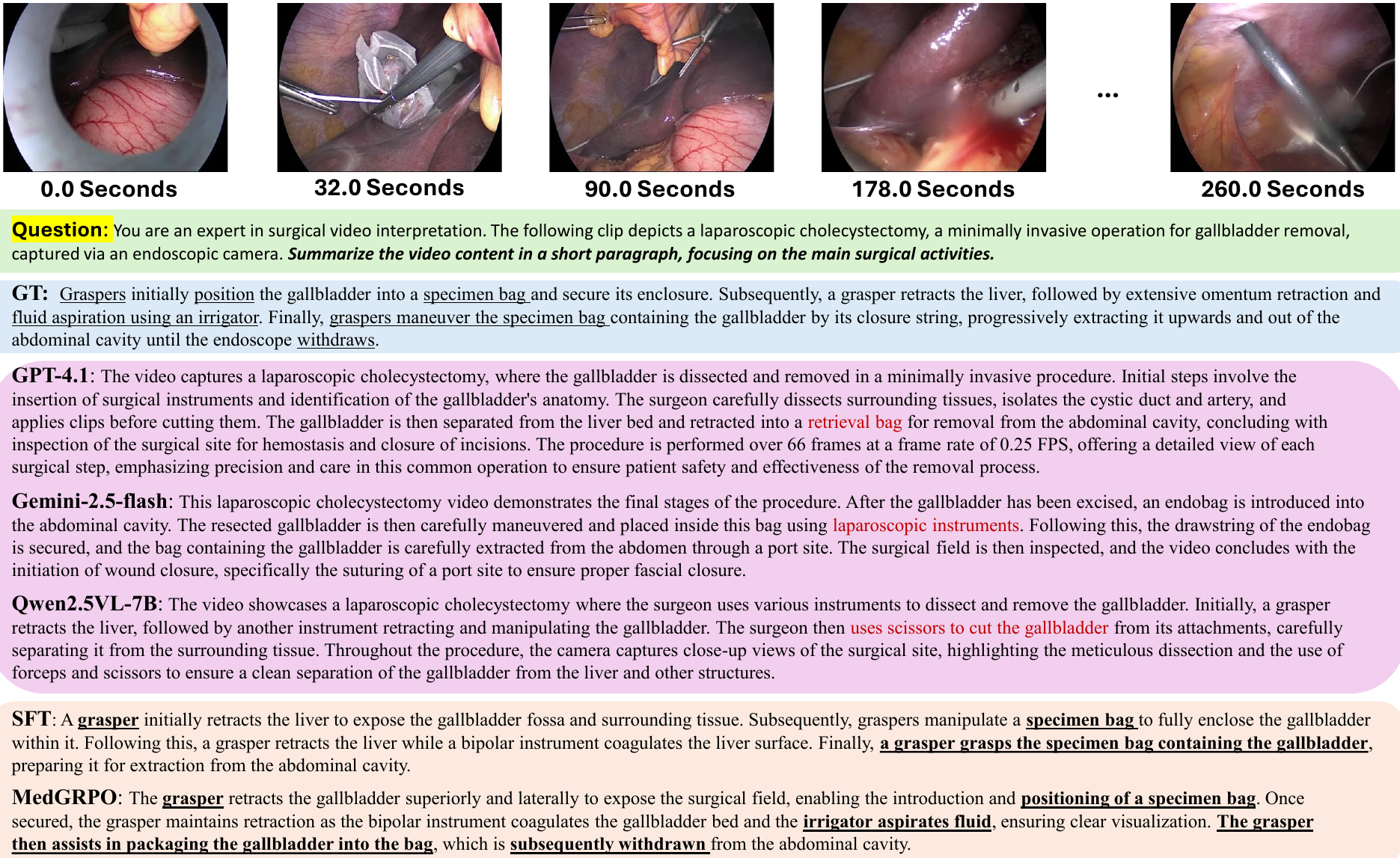}
\caption{Qualitative examples on video summary.}
\label{fig:qualitative_3}
\end{figure*}

\section{Medical LLM Judge}
\label{sec:supp_llm_judge}

As described in \S3.4 of the main paper, we employ an LLM-as-a-judge approach using GPT-4.1 to evaluate caption quality through comparative assessment across five medical-domain-specific dimensions. Each dimension uses a 1--5 scale measuring how closely the generated caption matches the reference: \textbf{5} (very close match, minor phrasing differences), \textbf{4} (good match, minor omissions), \textbf{3} (partial match, notable omissions), \textbf{2} (significant differences, missing important information), \textbf{1} (very different, major errors or missing content). \S\ref{sec:supp_llm_judge:quality} details how the LLM Judge scores correlates with scores from board certified clinicians on these five dimensions. We describe these five dimensions in details in \S\ref{sec:supp_llm_judge:rubrics} with scoring rubrics.

\subsection{Quality Assurance}
\label{sec:supp_llm_judge:quality}
To establish the validity of our LLM-as-a-judge approach using GPT-4.1 to evaluate caption quality, we conduct a rigorous human study with \textbf{10 board-certified clinicians}. The clinicians were asked to score 30 samples across the same 5 clinical dimensions used by our LLM Judge (paper lines 404-410). Results show strong correlation: Pearson $r$=$0.977$, Cohen's Kappa =$0.817$, confirming our automated metric effectively proxies human clinical preference. We further show the correlation of the LLM Judge with clinicians across all the five evaluation dimensions ($R1$-$R5$) in Fig.~\ref{fig:clinician_LLM}. This highlights that the LLM Judge very closely agrees with clinicians in all of the evaluation dimensions.

\subsection{Detailed Rubrics}
\label{sec:supp_llm_judge:rubrics}
These five evaluation dimensions used by the LLM Judge and board certified clinicians to score captions is detailed in this section.
\paragraph{Medical Terminology Precision (R1).}
\textit{Definition}: Does the generated caption use the same medical terms as the reference?

\textit{Scoring Rubric:}
\begin{itemize}
    \item Score 5: medical terms match reference precisely (instruments, anatomy, actions)
    \item Score 4: most terms match reference, minor substitutions acceptable
    \item Score 3: some terms match reference, some generic or imprecise
    \item Score 2: many terms don't match reference, often generic
    \item Score 1: terms mostly don't match reference or are incorrect
\end{itemize}

\paragraph{Instrument and Anatomy Identification (R2).}
\textit{Definition}: Are the instruments and anatomical structures identified the same as in the reference?

\textit{Scoring Rubrics:}
\begin{itemize}
    \item Score 5: all instruments and anatomy match reference identifications
    \item Score 4: most instruments and anatomy match reference
    \item Score 3: some instruments and anatomy match reference, some missing
    \item Score 2: many instruments and anatomy don't match reference
    \item Score 1: instruments and anatomy mostly wrong or missing vs reference
\end{itemize}

\paragraph{Specificity vs Vagueness (R3).}
\textit{Definition}: Is the level of specificity/vagueness similar to the reference?

\textit{Scoring Rubrics:}
\begin{itemize}
    \item Score 5: specificity level matches reference (specific when reference is specific)
    \item Score 4: specificity level mostly matches reference
    \item Score 3: specificity level sometimes differs from reference
    \item Score 2: specificity level often differs from reference (too vague or too specific)
    \item Score 1: specificity level doesn't match reference at all
\end{itemize}

\paragraph{Clinical Procedure Context (R4).}
\textit{Definition}: Does the generated caption convey the same procedural understanding as the reference?

\textit{Scoring Rubrics:}
\begin{itemize}
    \item Score 5: procedural context matches reference (workflow, steps, purpose)
    \item Score 4: most procedural context matches reference
    \item Score 3: some procedural context matches reference, some missing
    \item Score 2: procedural context differs significantly from reference
    \item Score 1: procedural context mostly missing or wrong vs reference
\end{itemize}

\paragraph{Action and State Accuracy (R5).}
\textit{Definition}: Are the actions and states described the same as in the reference?

\textit{Scoring Rubrics:} 
\begin{itemize}
    \item Score 5: all actions and states match reference (active/idle, grasping/releasing, etc.)
    \item Score 4: most actions and states match reference
    \item Score 3: some actions and states match reference, some differ
    \item Score 2: many actions and states differ from reference
    \item Score 1: actions and states mostly wrong vs reference
\end{itemize}


\section{Additional Qualitative Results}
\label{sec:supp_qualitative}
\paragraph{Task Examples.} Figure \ref{fig:dataset_overview} (c) in the main paper shows three tasks (Skill Assessment, Region Captioning, Temporal Action Grounding) across nursing, laparoscopic and robotic surgery. Figure \ref{fig:more_task} provides five additional examples (Dense Video Captioning, Spatio-Temporal Grounding, Critical View Safety, Video Summary, Next Action Prediction) spanning nursing, laparoscopic and open surgery. Together, these examples showcase MedVidBench's coverage of 8 diverse tasks across 4 medical domains (nursing, robotic, laparoscopic and open surgery).

\paragraph{Sensitivity of Slope $k$:} Figure~\ref{fig:stg_k} shows reward distributions for STG across datasets. \textbf{w/o normalization} (left), distributions are dramatically imbalanced—easy datasets yield consistently higher rewards than hard ones, causing training collapse. \textbf{w/ normalization} (right), all datasets achieve balanced, centered distributions. Crucially, varying $k \in \{2, 3, 4\}$ produces nearly identical normalized distributions, showing our method is relatively insensitive to $k$.

\paragraph{Dense Video Captioning.}
Figure~\ref{fig:qualitative_2} shows qualitative comparisons for dense video captioning on a Penicillin Skin Testing procedure. MedGRPO demonstrates superior performance over off-the-shelf models and SFT baseline: (1) \textbf{action identification}—correctly identifies both key actions (disinfect skin at 0.0--9.0s, perform intradermal puncture at 19.0--27.0s) matching ground truth (GT: 0.0--5.0s and 13.0--34.0s), while GPT-4.1 completely misses both actions and invents non-existent ``prepare skin test solution'' and ``observe results'', Gemini-2.5-flash captures only partial disinfection (0.0--1.0s), and Qwen2.5VL-7B generates highly repetitive, non-specific descriptions across 23.0--58.0s without naming explicit actions; (2) \textbf{precise terminology}—uses specific clinical terms (``intradermal puncture'', ``volar aspect of forearm'', ``circular motions'', ``shallow angle'', ``epidermis'', ``visible wheal'') matching GT vocabulary, versus Gemini-2.5-flash's vague ``cleaning a skin-like surface'' and Qwen2.5VL-7B's generic ``holds a syringe...preparing for injection''; (3) \textbf{technical detail}—captures procedural specifics including ``alcohol swab'', ``air-dry'', ``needle at shallow angle into epidermis'', ``brief retraction'', and ``visible wheal formation'', closely aligning with GT descriptions; and (4) \textbf{temporal accuracy}—provides reasonable temporal boundaries with minor deviations (4s extension for disinfection, 6s shift for puncture), while SFT slightly underestimates disinfection duration (0.0--4.5s) and off-the-shelf models show severe temporal misalignment or excessive repetition.
\begin{figure*}[t]
\centering
\includegraphics[width=\linewidth]{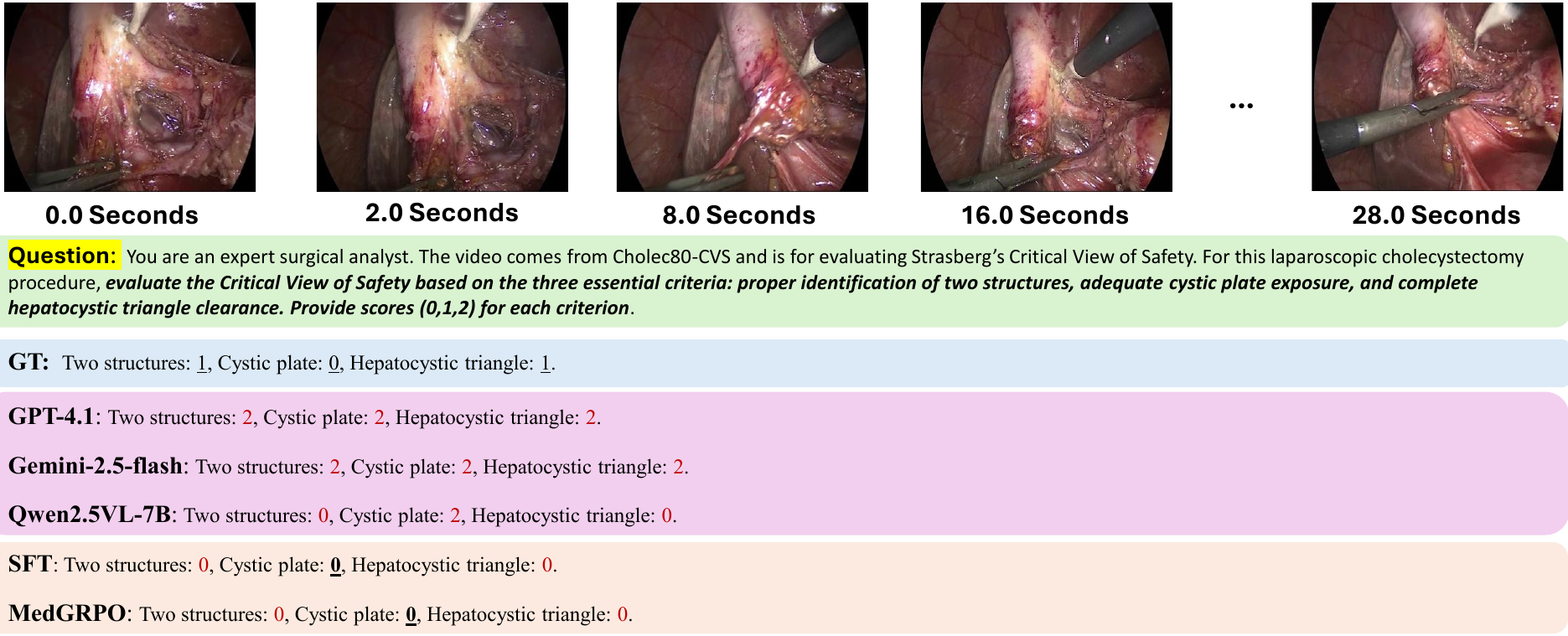}
\caption{Failure case examples on Critical View of Safety (CVS) assessment.}
\label{fig:failure}
\vspace{-1em}
\end{figure*}
\paragraph{Video Summary.}
Figure~\ref{fig:qualitative_3} presents qualitative comparisons for video summarization on a final part of laparoscopic cholecystectomy. MedGRPO produces a clinically accurate summary closely matching ground truth: grasper retracts gallbladder and positions specimen bag, bipolar instrument coagulates gallbladder bed, irrigator aspirates fluid for visualization, followed by specimen bag packaging and extraction. Key advantages: (1) \textbf{procedural accuracy}—correctly identifies the post-excision extraction phase matching GT (gallbladder positioning into specimen bag, liver retraction, omentum retraction, fluid aspiration, extraction), while GPT-4.1 hallucinates entire dissection workflow (``dissected and removed'', ``dissects surrounding tissues'', ``applies clips before cutting'') not present in video, and Qwen2.5VL-7B invents ``scissors cutting the gallbladder from attachments'' despite no cutting occurring; (2) \textbf{instrument identification}—accurately names specific instruments (grasper, bipolar instrument, irrigator) with their functions, whereas Gemini-2.5-flash uses generic terms (``laparoscopic instruments'') and GPT-4.1 describes non-existent ``clips''; (3) \textbf{workflow completeness}—captures the full sequence (positioning, liver/omentum retraction, fluid aspiration, coagulation, extraction) matching GT, while SFT omits critical irrigation and fluid aspiration steps entirely; and (4) \textbf{temporal focus}—correctly focuses on specimen removal phase, unlike GPT-4.1's fabricated early-stage dissection narrative spanning ``66 frames at 0.25 FPS''.

\paragraph{Failure Analysis.}
Figure~\ref{fig:failure} illustrates a representative failure case on Critical View of Safety (CVS) assessment from Cholec80-CVS, evaluating three criteria with scores 0--2 per criterion. Ground truth scores: two structures (1), cystic plate (0), hepatocystic triangle (1). Despite strong performance on other tasks, all models struggle with CVS scoring: \textbf{Scoring calibration issues}—MedGRPO and SFT both score conservatively (0, 0, 0), systematically underestimating versus ground truth, while GPT-4.1 and Gemini-2.5-flash consistently overestimate (2, 2, 2), and Qwen2.5VL-7B shows erratic scoring (0, 2, 0). No model correctly identifies the intermediate ground truth pattern (1, 0, 1), suggesting fundamental difficulty in calibrating to surgical assessment rubrics rather than model-specific failure. \textbf{Intermediate score challenge}—with only 3.8K CVS training samples, models struggle to distinguish between partial achievement (score 1) versus absent (0) or complete (2), defaulting to extreme scores. CVS requires nuanced anatomical judgment: partial structure identification, subtle tissue plane visualization, and incomplete clearance assessment. \textbf{Multi-criteria reasoning}—simultaneous evaluation of three interrelated surgical safety criteria demands integrated anatomical knowledge and spatial reasoning that current models lack. Future work should explore specialized scoring calibration mechanisms, confidence-aware predictions for ambiguous cases, and expanded training data for underrepresented surgical evaluation tasks.

\end{document}